\documentclass[lettersize,journal]{IEEEtran}
\usepackage{amsmath,amsfonts}
\usepackage{algorithmic}
\usepackage{algorithm}
\usepackage{array}
\usepackage[caption=false,font=normalsize,labelfont=sf,textfont=sf]{subfig}
\usepackage{textcomp}
\usepackage{stfloats}
\usepackage{url}
\usepackage{verbatim}
\usepackage{graphicx}
\usepackage{cite}
\usepackage{amsmath, amssymb, amsthm}
\newtheorem{lemma}{Lemma}            
\newtheorem{assumption}{Assumption}  
\usepackage{booktabs}
\usepackage{makecell}
\usepackage{multirow}
\usepackage{float} 

\usepackage{color}
\usepackage{colortbl}
\usepackage{xcolor}
\newcommand{\wenlj}[1]{#1}

\definecolor{CUHKmiddle}{RGB}{144,44,144}
\usepackage{tcolorbox}
\tcbuselibrary{skins,breakable}
    {\endtcolorbox}

\begin{document}

\title{Multimodal Information Interactions Bottleneck for Multimodal Self-Supervised Learning}

\author{Liangjian Wen, Jiangtao Zheng, Jiang Duan$^*$, Qun Dai, Yong Dai, Jianzhuang Liu, Jun Wang, Shu-Dong Huang, Xinwang Liu, Zenglin Xu, Zhao Kang$^*$

\thanks{
L. Wen, J. Zheng, J. Duan, and Q. Dai are with the School of Computing and Artificial Intelligence,
Southwestern University of Finance and Economics, Chengdu, China.
J. Wang is with the School of Management Science and Engineering, Southwestern University of Finance and Economics, Chengdu, China.
L. Wen is also with the Artificial Intelligence and Digital Finance Key Laboratory of Sichuan Province (e-mail: wlj6816@gmail.com).
J. Duan is also with Chengdu Everimaging Science and Technology Co., Ltd.
}

\thanks{
Y. Dai is with the X-Humanoid Research Institute, Beijing, China.
}

\thanks{
J. Liu is with the Shenzhen Institutes of Advanced Technology,
Chinese Academy of Sciences, Shenzhen, China (e-mail: jz.liu@siat.ac.cn).
}
\thanks{
S. Huang is with the School of Artificial Intelligence, Sichuan University, Chengdu, China.
}
\thanks{X. Liu is with School of Computer, National University of Defense Technology, Changsha, 410073.}
\thanks{
Z. Xu is with the Artificial Intelligence Innovation and Incubation Institute,
Fudan University and also with the Shanghai Academy of AI for Science,
Shanghai, China.
}
\thanks{
Z. Kang is with the University of Electronic Science and Technology of China,
Chengdu, China.
}
\thanks{$^*$Jiang Duan and Zhao Kang are the corresponding authors (e-mail: duanj\_t@swufe.edu.cn; zkang@uestc.edu.cn).}
\thanks{This work was supported by the Major Science and Technology Special Project of the Sichuan Provincial Department of Science and Technology (Grant No. 2024ZDZX0002), the National Natural Science Foundation of China (No. U24A20323), Fundamental Research Funds for the Central Universities (JBK202511081), the Sichuan Provincial Innovation Group Project (Grant No. 2024NSFTD0054), Natural Science Foundation of China under Grant No. 62502397, National Natural Science Foundation of China (NSFC) (72471197), and the Sichuan Provincial Philosophy and Social Science Fund (Grant No. SCJJ25ND091).}
}

\markboth{IEEE Transactions on Pattern Analysis and Machine Intelligence}%
{Wen \MakeLowercase{\textit{et al.}}: Multimodal Information Interactions Bottleneck for Multimodal Self-Supervised Learning}


\maketitle

\begin{abstract}
Multimodal self-supervised learning is challenged by heterogeneity, unequal informativeness, and complex information interactions across multimodal data sources such as images, text, audio, and tabular data.
While contrastive methods effectively align shared semantics, they typically rely on strong multi-view redundancy assumptions, overlooking modality-specific and synergistic information, thus leading to suboptimal representations.
We propose the Multimodal Information Interactions Bottleneck (MIIB), a principled framework grounded in Partial Information Decomposition that explicitly \wenlj{encodes redundant, unique, and synergistic information within multimodal interactions}.
MIIB employs a dual-path model architecture for modality-specific encoding and multimodal fusion, optimized using \wenlj{surrogate} tractable entropy-based objectives that enforce minimal sufficient representations without requiring explicit mutual information estimation.
By leveraging variational bounds and score-based entropy estimation, MIIB efficiently captures complex cross-modal interactions while suppressing superfluous information.
Experiments across diverse benchmarks demonstrate consistent improvements over state-of-the-art multimodal self-supervised methods.
Code is available at \url{https://github.com/wanglingzaishi/MIIB}.
\end{abstract}


\section{Introduction}
\IEEEPARstart{M}{ultimodal} data, which encompasses diverse sources such as images, text, audio, and tabular structures, has become ubiquitous in real-world applications, including healthcare diagnostics~\cite{soenksen2022integrated,tu2024towards}, autonomous driving~\cite{caesar2020nuscenes,hu2023planning}, multimedia retrieval~\cite{radford2021learning,alayrac2022flamingo}, and human-computer interaction~\cite{baltruvsaitis2018multimodal,liu2023visual}.
These data exhibit key challenges: (1) heterogeneity in distributions (e.g., differing statistical properties between sparse textual data and dense visual data); (2) varying levels of informativeness across modalities (e.g., one modality may provide rich semantic details while another offers only supplementary context); and (3) complex information interactions among modalities (e.g., a harmless cartoon image paired with a simple caption, which together subtly endorses extremist ideology, though neither does so alone).
Although self-supervised learning methods, such as SimCLR~\cite{chen2020simple} and MAE~\cite{he2022masked}, have achieved remarkable success in unimodal domains, a critical question remains: how can multimodal self-supervised learning effectively overcome the inherent challenges of multimodal data?

Some works, such as CLIP~\cite{radford2021learning} and ALIGN~\cite{jia2021scaling}, try to align representations across modalities by treating them as multi-view augmentations, thereby capturing shared semantics.
Although these methods are effective for tasks like zero-shot classification and cross-modal retrieval, they often depend on a restrictive assumption known as the \textit{multi-view redundancy assumption}~\cite{DBLP:conf/colt/SridharanK08, DBLP:conf/iclr/Tsai0SM21, wen2024mveb}. This assumption states that each modality is (approximately) sufficient on its own for predicting downstream tasks and contains the same task-relevant information as the others.
However, this assumption originates from traditional multi-view learning and has limitations in real-world multimodal scenarios, where many tasks involve only minimal shared information across modalities.
To address this limitation, FactorCL~\cite{liang2023factorized} explicitly decomposes representations into shared and unique components, allowing the model to learn the unique information from each modality beyond just the multi-view redundancy.
In addition, \cite{dufumier2024align} extends the multi-view redundancy paradigm from contrastive learning to better suit multimodal scenarios. However, multimodal representation learning involves multiple modalities, introducing complex information interactions that can undermine the efficacy of straightforward contrastive approaches, as these approaches mainly focus on learning invariant features through data augmentations across multiple views.

To understand multimodal interactions, we can use Partial Information Decomposition (PID)~\cite{williams2010nonnegative,bertschinger2014quantifying,wen2026dependency}.
In the PID framework, these interactions can be broken down into three main components:
Redundancy (R): This represents information that is shared across multiple modalities. For example, in an image-text pair describing a scene, both modalities may convey the same core concepts (e.g., object identity), leading to overlapping signals. While redundancy can reinforce robustness in learning, excessive reliance on it in contrastive objectives may cause overfitting to shared artifacts, reducing the model's ability to generalize.
Uniqueness (U): This encompasses information specific to a single modality, which is not available in others. For instance, spatial details in images or syntactic nuances in text are unique contributions. Modality-specific noise, such as visual clutter or linguistic ambiguities, often falls under this category, potentially introducing distortions that misalign cross-modal representations if not properly mitigated.
Synergy (S): This denotes complementary information that emerges only through the combination of modalities, providing insights beyond what individual sources offer. An example is inferring sentiment from audio tone combined with textual content, where the joint analysis reveals nuances absent in isolation. 

\begin{figure}[t!]
    \centering
    \includegraphics[width=\columnwidth, keepaspectratio]{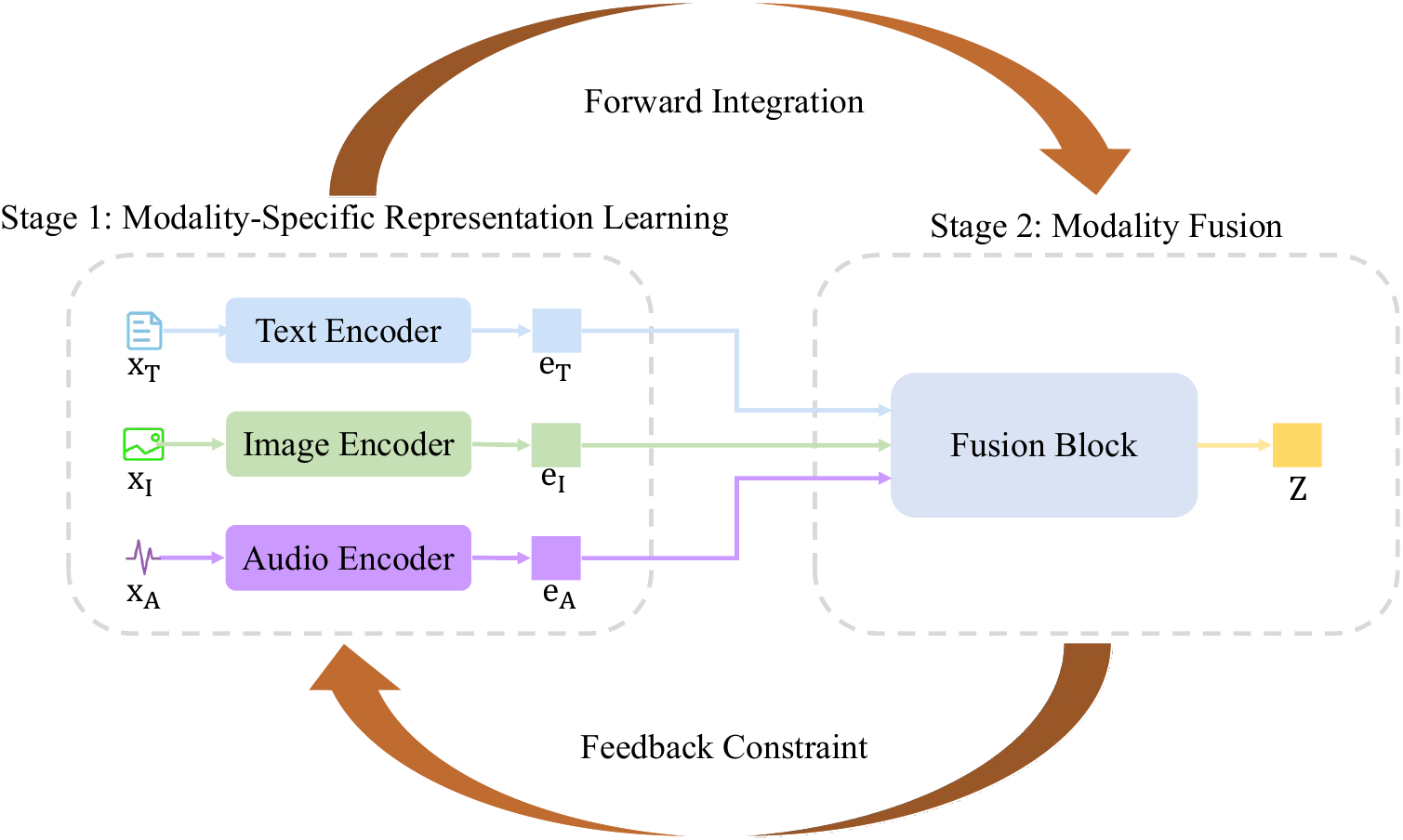}
    
    \caption{The process framework for multimodal representation learning organizes modality-specific learning and fusion as a sequential yet interdependent process.}
    \label{fig:introduction}
\end{figure}


The information-theoretic PID framework aligns with the broader structure of multimodal representation learning. As shown in Fig.~\ref{fig:introduction}, multimodal representation learning comprises two primary components: individual modality-specific representation learning and modality fusion. 
The former focuses on extracting latent features from each modality independently, leveraging modality-tailored encoders to capture intrinsic patterns and reduce dimensionality while preserving modality-unique and redundant information. 
Due to the heterogeneity in distributions of multimodal data and varying levels of informativeness across modalities, it is challenging to learn effective individual modality-specific representations.
The latter integrates these modality-specific representations into a unified multimodal embedding, often through mechanisms to exploit cross-modal synergies. Capturing these complex interactions effectively is essential for multimodal representation learning, enabling models to generalize across varied tasks.

The process of modality-specific representation learning and modality fusion in multimodal systems is inherently sequential and interdependent. Specifically, modality-specific representation learning acts as a foundational preprocessing stage, wherein each individual modality—such as text, images, or audio—is processed independently to extract, isolate, and refine its unique signals. This step mitigates noise, enhances feature relevance, and generates high-quality embeddings that serve as reliable inputs for subsequent fusion. In contrast, the modality fusion phase integrates these preprocessed representations by aligning them across modalities and combining complementary information, thereby enriching the overall representational quality. This integration enables the model to capture holistic multimodal patterns, leading to superior performance on downstream tasks that demand comprehensive cross-modal understanding. However, the effectiveness of the modality fusion process can, in turn, impose constraints on modality-specific representation learning; for instance, suboptimal fusion outcomes—such as misalignment or overemphasis on certain modalities—may necessitate iterative refinements in the individual modality encoders, creating a feedback loop that influences the entire learning pipeline.

In the multimodal representation learning process, if interactions between certain modalities are not adequately captured—due to factors like insufficient training data or architectural limitations—this deficiency propagates through the interdependent nature of modality-specific representation learning and modality fusion. These shortcomings can result in mutual degradation: suboptimal modality-specific features impede effective fusion, while imperfect fusion exacerbates errors in individual representations. This ultimately hinders the learning of effective multimodal representations.
To address these challenges, building on this foundation, we propose the Multimodal Information Interactions Bottleneck (MIIB), a novel self-supervised framework that enforces minimal sufficient representations at both modality-specific and fusion levels. MIIB introduces a dual-path architecture with specialized Transformer blocks for integrated fusion, modality-specific processing, and synergistic extraction, optimized via a simplified \wenlj{surrogate } loss that balances minimality, sufficiency, and disentanglement without relying on intractable mutual information estimators. By incorporating variational bounds and score-based entropy estimation\cite{wen2024mveb}, MIIB efficiently captures complex interactions while reducing superfluous information, leading to more generalizable multimodal representations.

Our contributions are threefold: (1) We formalize the multimodal information interactions bottleneck from an information-theoretic perspective, highlighting how interdependent modality-specific learning and fusion exacerbate inefficiencies in existing methods. (2) We introduce MIIB, a principled framework that promotes effective multimodal integration through targeted losses for modality-specific, fusion, and synergistic representations. (3) Through extensive experiments on benchmarks, we demonstrate that MIIB outperforms state-of-the-art multimodal self-supervised models.

\section{Related Work}

\paragraph{Multimodal Learning} Multimodal learning aims to connect and integrate information from diverse data sources---such as text, images, audio, tactile, and tabular data---to enhance the understanding of cross-modal information~\cite{baltruvsaitis2018multimodal,akkus2023multimodal,guo2026webcogreasoner}. Early research focused on training separate encoders and relying on conventional fusion techniques like feature concatenation~\cite{d2015review} or modality-specific prediction averaging~\cite{friedman2008predictive} to aggregate information across modalities. With the development of Transformers and ViTs, the introduction of dynamic cross-modal attention mechanisms~\cite{tsai2019multimodal,xu2023multimodal} has enabled models to more effectively capture and model complex inter-modal interactions. Contemporary methodologies typically adhere to a two-stage framework: initially utilizing specialized encoders to extract features from specific modalities, and subsequently projecting these representations into a unified embedding space~\cite{dufumier2024align,baltruvsaitis2018multimodal} for alignment or fusion. This paradigm has been widely adopted in multimodal representation learning~\cite{barnum2020benefits,radford2021learning}, cross-modal alignment~\cite{li2022clip,hendricks2021decoupling}, multimodal clustering~\cite{lou2026granular,hu2026trust}, and generative modeling~\cite{ahuja2020style,ramesh2021zero}.
\paragraph{Self-Supervised Multimodal Representation Learning} Self-supervised learning aims to generate supervision signals from the inherent structure of data~\cite{balestriero2023cookbook,DBLP:conf/iclr/WenZHZX20}. In the multimodal domain, existing methodologies are widely organized into two representative paradigms. The first is the generative paradigm, which utilizes proxy tasks—such as cross-modal reconstruction~\cite{ramesh2021zero} or masked prediction~\cite{bachmann2022multimae}—to restore masked or absent inputs.
Alternatively, the contrastive learning framework~\cite{girdhar2023imagebind,radford2021learning} represents another extensively adopted strategy. Methods in this category leverage distinct modalities to formulate semantically complementary views, constructing positive pairs through data augmentation~\cite{chen2020simple}. They then employ contrastive objectives (e.g., InfoNCE) to minimize the distance between positive pairs while maximizing the separation between negative pairs, thereby aligning the semantic features of diverse modalities within the representation space~\cite{wang2022rethinking}.
\wenlj{A parallel line of work approaches common information through an identifiability lens, seeking to provably recover and disentangle the latent factors that are shared across views/tasks from those that are task-unique\cite{zhang2024discovering, kleinman2023gacs, achille2018emergence}.
MIIB is orthogonal: instead of disentangling shared/unique factors, it encodes task-relevant common information into one sufficiency–minimal bottleneck, with no identifiability assumptions.}

\paragraph{Multimodal Contrastive Learning}
While conventional contrastive learning frameworks~\cite{radford2021learning,jia2021scaling} excel in cross-modal alignment, they implicitly rely on the multiview redundancy assumption~\cite{tian2020makes,lin2022dual}, positing that shared information suffices for downstream tasks. Consequently, these methods predominantly emphasize shared representations, neglecting to explicitly model uniqueness and synergistic interactions.
To circumvent this limitation, FactorCL~\cite{liang2023factorized} attempts to explicitly disentangle shared and unique information, yet it relies on restrictive augmentation assumptions and fails to model synergistic interactions. 
Subsequently, CoMM~\cite{dufumier2024align} employs multimodal data augmentation and information-theoretic loss functions, aiming to capture redundancy, uniqueness, and synergy within a shared representation space. 
However, while CoMM theoretically encompasses all three types of interactions, its mechanism for modeling complex synergistic interactions leaves room for further optimization. 
Consequently, designing effective mechanisms to enhance the capture of synergistic information remains a critical challenge in the field~\cite{wen2025infmasking}. Beyond contrastive alignment, HRIL explores higher-order tensor modeling for multimodal synergy~\cite{dai2026hril}, while MSCR jointly addresses modality utilization balance and synergistic information discovery~\cite{chen2026mscr}.

\wenlj{\textbf{Positioning against FactorCL and CoMM}.
MIIB differs in three respects. (i) Instead of regularizing one fused space, it models multimodal learning as two interdependent processes: modality-specific encoding and fusion, and imposes a minimal-sufficiency bottleneck at both levels (Fig. 1), so the two stages constrain each other. (ii) It gives synergy a dedicated stream from a separate Synergy Transformer and supervises it with the fused cross-view representations: each synergy representation is trained to predict the other view's fused representation, which can be satisfied only by information jointly carried across modalities — so synergy is learned, not merely encouraged to differ from the unimodal streams. (iii) The objective is reduced to a finite hypersphere entropy-and-alignment regularizer estimated by a score-based von Mises–Fisher estimator, avoiding the direct mutual-information estimation that both FactorCL and CoMM rely on.}

\section{Preliminary }
\subsection{Minimal Sufficient Representation}

Under the information bottleneck principle~\cite{tishby2000information,hu2024survey}, a representation \( z \) is considered sufficient for a target variable \( Y \) if it preserves all relevant information in the input \( X \) necessary for predicting \( Y \). Among all sufficient representations, a minimal sufficient representation is one that achieves this sufficiency while discarding as much superfluous (i.e., task-irrelevant) information from \( X \) as possible. This concept, originally defined for supervised tasks, has been extended to self-supervised multi-view learning.

Under the multi-view assumption, where two views \( v_1 \) and \( v_2 \) share task-relevant semantics, a representation \( z_1 \) is sufficient if it captures the shared information \( I(v_1; v_2) \). It is minimal if it eliminates view-specific superfluous information by minimizing the conditional mutual information \( I(z_1; v_1 | v_2) \)~\cite{federici2020learning}.

\section{Contrastive Multimodal Learning From an Information Perspective}
Let $  X_1, X_2, \dots, X_n  $ denote random variables, each embodying data from a unique modality (such as images, text, audio, or tabular structures), accompanied by a downstream objective represented by $  Y  $.
The goal is to construct a robust multimodal latent representation $  Z_{\theta} = f_{\theta}(X)  $, where $  X = (X_1, \dots, X_n)  $ and $  \theta  $ signifies the parameters of the multimodal encoder, such that $  Z_{\theta}  $ qualifies as effective by encapsulating task-relevant details from $  Y  $ while preserving mutual information: $  I(Z; Y) = I(X; Y)  $.

In self-supervised learning, the target variable $  Y  $ is left undefined throughout the representation learning stage, thereby introducing a distinctive obstacle.
To mitigate this issue, Dufumier et al. \cite{dufumier2024align} generalize the multiview redundancy paradigm from contrastive learning to encompass multimodal scenarios:

\begin{assumption}[\textit{Minimal Label-Preserving Multimodal Augmentations}]
\label{assumption: label-preserving}
There exists a collection $  \mathcal{T}^*  $ of multimodal operators such that, for every $  t \in \mathcal{T}^*  $ and the transformed instance $  X' = t(X)  $, the equality $  I(X; X') = I(X; Y)  $ is satisfied, thereby maintaining the informational relevance to the label $  Y  $.
\end{assumption}

Assumption~\ref{assumption: label-preserving} holds validity in the context of multimodal representation learning. This assumption facilitates an expanded spectrum of data augmentations, unconstrained by the specific ensemble $  \mathcal{T}_c^{\star} = \{ t(X) = (t_1(X_1), t_2(X_2), \dots, t_n(X_n)) \}  $.
Within this framework, $  X'  $ assumes a function comparable to supervisory signals in labeled training paradigms.
We introduce $  Z_{\theta}' = f_{\theta}(X')  $, with $  X' = t(X)  $ and $  t \in \mathcal{T}  $ denoting a multimodal transformation.
Leveraging the data processing inequality applied to the Markov chain $  X \to X' \to Z_{\theta}'  $, we derive the ensuing bounds on mutual information:
\begin{align}
I(Z_\theta; Z_\theta') \leq I(Z_\theta; X') \leq I(X; X').
\end{align}
This inequality underpins the subsequent lemma:
\begin{lemma}\label{lemma: Ixxp}
In the process of optimizing $  f_\theta  $ to maximize $  I(Z_\theta; Z_\theta')  $, assuming $  f_\theta  $ exhibits adequate representational capacity, the optimal parameters $  \theta^\star  $ yield $  I(Z_{\theta^\star}; Z_{\theta^\star}') = I(Z_{\theta^\star}; X') = I(X; X')  $.
\end{lemma}
From Lemma~\ref{lemma: Ixxp}, we observe that maximizing the mutual information $  I(Z_\theta; Z_\theta')  $ effectively aligns the representations to capture the shared information between $  X  $ and $  X'  $, which, under Assumption~\ref{assumption: label-preserving}, preserves the task-relevant information for $ Y $.

Although contrastive multimodal learning can, in theory, yield effective representations, maximizing the mutual information $  I(Z_\theta; Z_\theta')  $ in practice often leads to suboptimal performance, particularly in real-world applications involving diverse datasets. This arises because contrastive multimodal learning primarily emphasizes learning invariant features under data augmentations from multiple views.

Multimodal representation learning involves integrating information from diverse sources such as images, text, and audio. This fundamental difference arises primarily from the intricate interactions among multiple modalities, where each modality may contribute unique yet complementary information that must be harmoniously fused to form a cohesive representation. 
As a result, multimodal approaches require sophisticated mechanisms to capture the complex interactions among multiple modalities.

\section{Method}

\subsection{ Multimodal  Information Interactions 
Bottleneck  }
\textbf{Multimodal   Interactions}. Multimodal interactions refer to the process of extracting and integrating information from diverse data sources, such as text, images, audio, or tabular data, to form a cohesive representation for downstream tasks. To understand these multimodal interactions, we can leverage Partial Information Decomposition (PID)~\cite{williams2010nonnegative,bertschinger2014quantifying}, a theoretical framework that dissects the mutual information among multiple variables into non-overlapping components. 
Under the PID framework, these interactions can be decomposed into three primary components:
Redundancy (R): This represents information that is shared across multiple modalities.
Uniqueness (U): This encompasses information specific to a single modality that is not available in others. Synergy (S): This denotes complementary information that emerges only through the combination of modalities, providing insights beyond what individual sources offer. 

For analytical tractability, we focus on the case of $  n = 2  $. Consider two modalities $  X_1  $ and $  X_2  $, along with a task $  Y  $. However, it can be directly extended to $n$ modalities.
The mutual information $  I(X_1, X_2; Y)  $ measures the total task-relevant information from both modalities combined. Using partial information decomposition (PID), this breaks down as:
\begin{align}
I(X_1, X_2; Y) = R + S + U_1 + U_2, 
\end{align}
where $  R  $ is redundant information shared by $  X_1  $ and $  X_2  $, $  S  $ is synergistic information arising only from their integration, and $  U_1  $ and $  U_2  $ are unique contributions from $  X_1  $ and $  X_2  $, respectively.
This breakdown aligns with consistency equations from the chain rule of mutual information:
\begin{align}
 &I(X_1; Y) = R + U_1, \quad I(X_2; Y) = R + U_2, \nonumber \\
 &\quad I(X_1; X_2; Y) = R - S. 
\end{align}

\begin{figure*}[t!]
    \centering
    
    \includegraphics[width=\textwidth, keepaspectratio]{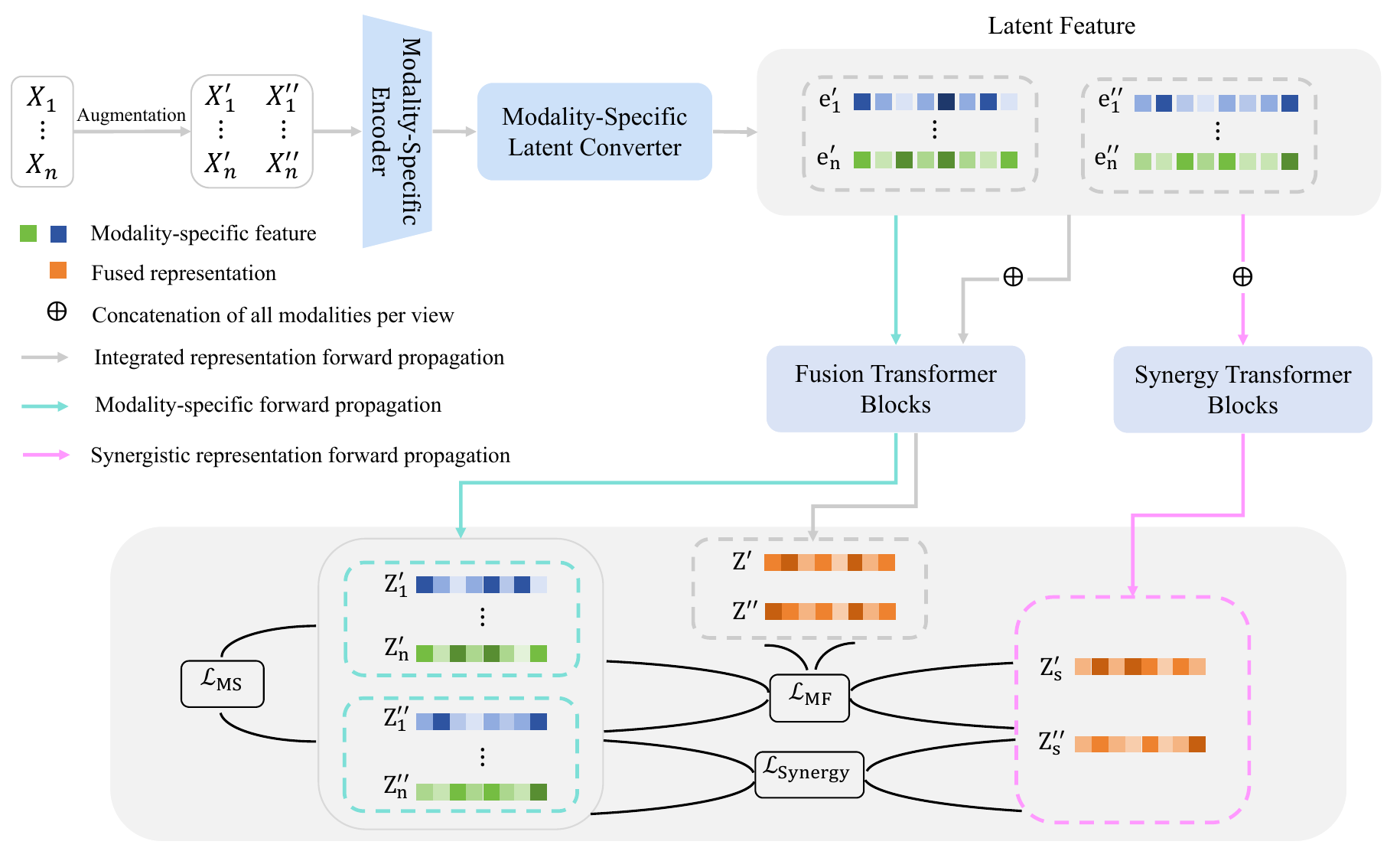}
    
    \caption{The overall architecture of MIIB. Given $n$ input modalities $X = (X_1, \dots, X_n)$, we first augment them to generate two views, $X'$ and $X''$. These views are encoded independently by modality-specific encoders and latent converters to extract latent features $e'_1 \dots e'_n$ and $e''_1 \dots e''_n$, \wenlj{which provide unified embeddings for subsequent multimodal interaction.}
    These features are then processed by two parallel Transformer streams. 
    The Fusion Transformer processes concatenated multimodal features to produce fused representations $Z', Z''$ while processing individual modality features to yield modality-specific representations $Z'_1 \dots Z'_n$ and $Z''_1 \dots Z''_n$. 
    \wenlj{The Synergy Transformer operates on the concatenated multimodal features to model cross-modal interactions, producing synergistic representations $Z'_s$ and $Z''_s$.}
    Finally, these representations are used to compute the modality-specific loss ($\mathcal{L}_{\text{MS}}$), modality fusion loss ($\mathcal{L}_{\text{MF}}$), and synergistic loss ($\mathcal{L}_{\text{Synergy}}$).}
    \label{fig:model}
\end{figure*}

We propose the Multimodal Information Interactions Bottleneck (MIIB), which preserves modality independence (to maintain the integrity of unique signals) while fostering collaborative integration (to leverage synergies across modalities). 
The architecture of MIIB is depicted in Fig. \ref{fig:model}, comprising two key phases: encoding latent features tailored to each modality and integrating multimodal features using a Transformer.
Starting with an input collection of $  n  $ modalities denoted as $  X = (X_1, X_2, \dots, X_n)  $, we derive augmented versions $  X'  $ and $  X''  $ via data augmentation. These are then fed into encoders specific to each modality, allowing independent extraction of latent representations  $  e'_1, e'_2, \dots, e'_n  $ and $  e''_1, e''_2, \dots, e''_n  $. As illustrated in Fig.\ref{fig:model}, the resulting features from these modalities undergo parallel processing in three separate paths:
(1) Features from all modalities are combined via concatenation and passed through a Transformer block to generate integrated representations $  Z'  $ and $  Z''  $; (2) Features from individual modalities are routed separately into Transformer blocks, resulting in modality-specific outputs $  Z'_1, Z'_2, \dots, Z'_n  $ and $  Z''_1, Z''_2, \dots, Z''_n  $;
(3) Features from all modalities are combined via concatenation and passed through another Transformer block to generate synergistic representations $  Z_{s}^{'}  $ and $ Z_{s}^{''} $. 

In the context of multimodal representation learning, we introduce the information bottleneck principle during the modality-specific representation learning phase to derive minimal sufficient representations for each modality, denoted as $  Z'_1, Z'_2, \dots, Z'_n  $ and $  Z''_1, Z''_2, \dots, Z''_n  $. Specifically, each modality feature $  Z'_i  $ incorporates a shared component $  R  $ along with unique modality-specific information $  U_i  $, thereby ensuring representational sufficiency while minimizing superfluous details. This approach yields robust, high-quality embeddings that serve as reliable inputs for subsequent fusion processes. To this end, we derive the following optimization objective using the relaxed Lagrangian multiplier method:
\begin{align}\label{eq:ms_loss_view1}
\mathcal{L}'_{MS}\left(\theta\right)=\sum_{i=1}^{n}\left[\lambda_{\text{MS}} I_\theta\left(Z'_i ; X'_i \mid X''\right)- I_\theta\left(Z'_i ; X''\right)\right],
\end{align}
where $  \theta  $ denotes the parameters of the multimodal model, and $  n  $ represents the number of modalities.
Based on Assumption~\ref{assumption: label-preserving}, \wenlj{for $Z'_i$, the input view is its own $X'_i$ and the supervisory (label-surrogate) view is the other view $X''$ under the Markov assumption \((X'_i \to Z'_i, \text{target } X'')\). } Consequently, maximizing $  I_\theta(Z'_i; X'')  $ facilitates the extraction of the relevant component $  R  $ alongside $  U_i  $, thereby ensuring sufficiency. Moreover, superfluous information can be identified through the conditional mutual information $  I_\theta\left(Z'_i; X'_i | X''\right)$. Reducing this measure effectively removes such superfluous information.
Considering the modality-specific representation $  Z''_i  $ and the self-supervised signal $  X'  $ \wenlj{under the Markov assumption \((X''_i \to Z''_i, \text{target } X')\)}, we can derive a similar optimization objective as follows:
\begin{align}\label{eq:ms_loss_view2}
\mathcal{L}''_{MS}\left(\theta\right)=\sum_{i=1}^{n}\left[\lambda_{\text{MS}} I_\theta\left(Z''_i ; X''_i \mid X'\right) - I_\theta\left(Z''_i ; X'\right)\right].
\end{align}

During the modality fusion process, we aim for the multimodal representations $  Z'  $ and $  Z''  $ to capture the task-relevant information $  R + S + \sum_{i=1}^n U_i  $. To facilitate multimodal information interaction under Partial Information Decomposition (PID), we introduce three types of information losses to promote this interaction.

First, $  Z'  $ and $  Z''  $ serve as mutual self-supervision signals. According to Lemma~\ref{lemma: Ixxp}, maximizing the mutual information $  I(Z_\theta; Z_\theta')  $ enables the learning of task-relevant information $  R + S + \sum_{i=1}^n U_i  $ for the target $  Y  $.

Second, the individual modality representations $  Z'_1, Z'_2, \dots, Z'_n  $ and $  Z''_1, Z''_2, \dots, Z''_n  $ act as self-supervision signals with respect to $  Z''  $ and $  Z'  $, respectively. Maximizing the mutual information $  \sum_{i=1}^n \left[ I_\theta(Z'_i; Z'') + I_\theta(Z''_i; Z') \right]  $ captures the shared information $  R + U_i  $ between each modality feature and the fused multimodal representation.

Third, the synergistic representations $  Z_s'  $ and $  Z_s''  $ serve as self-supervision signals with respect to $  Z''  $ and $  Z'  $, respectively. Maximizing the mutual information $  I(Z_s'; Z'') + I(Z_s''; Z')  $ learns the shared synergistic information $ S $.

To enhance the effectiveness of the modality fusion process, we minimize the conditional mutual information $  I_\theta(Z'; X' \mid X'')  $ and $  I_\theta(Z''; X'' \mid X')  $, thereby reducing superfluous information and facilitating minimal sufficient representation learning for $  Z''  $ and $  Z'  $. 
In summary, we derive the modality fusion loss as follows:
\begin{align}\label{eq:mf_loss}
\mathcal{L}_{MF}\left(\theta\right)= & \lambda_{\text{MF}} [I_\theta(Z'; X' \mid X'')+I_\theta(Z''; X'' \mid X') ]\nonumber\\
&-I_\theta(Z'; Z'') -\sum_{i=1}^n \left[ I_\theta(Z'_i; Z'') + I_\theta(Z''_i; Z')\right]\nonumber\\&- I_\theta(Z_s'; Z'') - I_\theta(Z_s''; Z')
\end{align}

To enhance synergistic interactions in our multimodal fusion framework, we aim to enforce minimal sufficient synergistic representations $  Z_s'  $ and $  Z_s''  $. These representations capture only the essential collaborative information shared between modalities, while minimizing redundancy. This approach ensures efficient integration without overwhelming the model with irrelevant details.
First, to reduce superfluous information in these representations, we derive an optimization objective based on conditional mutual information. Specifically, we minimize the dependency of $  Z_s'  $ on $  X'  $ given $  X''  $, and vice versa, which encourages $  Z_s'  $ and $  Z_s''  $ to focus solely on synergistic aspects:
\begin{align}
I_\theta\left(Z_{s}^{'}; X' | X''\right) + I_\theta\left(Z_{s}^{''}; X'' | X'\right).
\end{align}
Next,  we introduce a disentangled information loss to further enable the Transformer to extract pure synergistic information. This loss promotes separation by penalizing mutual information between individual modality-specific features ($  Z'_i  $ and $  Z''_i  $) and the opposing synergistic representations ($  Z_s''  $ and $  Z_s'  $), across $  n  $ elements
\begin{align}
\sum_{i=1}^n \left[ I_\theta(Z'_i; Z_s'') + I_\theta(Z''_i; Z_s')\right ].
\end{align}
By combining these components, we formulate the overall optimization loss for the synergistic representations in multimodal fusion ($  \mathcal{L}_{\text{synergy}}  $).
\begin{align}
\mathcal{L}_{\text{synergy}}\left(\theta\right)&=\lambda_{\text{ss}}[I_\theta\left(Z_{s}^{'}; X' | X''\right) + I_\theta\left(Z_{s}^{''}; X'' | X'\right)] \nonumber\\&+ \sum_{i=1}^n \left[ I_\theta(Z'_i; Z_s'') + I_\theta( Z''_i; Z_s')\right ].
\end{align}

\wenlj{Consider the term $  -I(Z_s';Z'')-I(Z_s'';Z')  $ that appears in the mutual-fusion loss $  \mathcal{L}_{MF}  $. Minimizing this term is mathematically equivalent to maximizing the sum of mutual informations
$I(Z_s';Z'') + I(Z_s'';Z').$
Under this objective, the synergistic representation $  Z_s'  $ and the modality-fused features $  Z''  $ of the opposite view mutually serve as self-supervision signals (and symmetrically between $  Z_s''  $ and $  Z'  $). Consequently, each synergy stream is encouraged to align with the full fused representation produced by the other view. Because $  Z''  $ aggregates information from both modalities of the opposite view, a high value of $  I(Z_s';Z'')  $ can be attained only when $  Z_s'  $ encodes information that is jointly carried across the two views. This cross-view mutual-information term therefore supplies the precise learning signal that enables the model to capture the shared synergistic information $  S  $.}
Hence, by incorporating the terms $  -I(Z_s'; Z'') - I(Z_s''; Z')  $ from $  \mathcal{L}_{MF}  $, this optimization objective promotes the learning of minimality, sufficiency, and disentanglement.

In summary, the MIIB framework is constructed by integrating three core optimization objectives to address the challenges in multimodal representation learning. The modality-specific loss $  \mathcal{L}_{MS} = \mathcal{L}'_{MS} + \mathcal{L}''_{MS}  $ ensures minimal sufficient representations for each individual modality by minimizing superfluous information while maximizing task-relevant components. The modality fusion loss $  \mathcal{L}_{MF}  $ promotes effective integration across modalities through mutual self-supervision, capturing shared, unique, and synergistic information while reducing superfluous information. Finally, the synergy loss $  \mathcal{L}_{\text{synergy}}  $ enforces disentangled and minimal sufficient synergistic representations, enhancing collaborative interactions without introducing irrelevant details. Together, these losses form the comprehensive MIIB objective, optimized as follows:
\begin{align}
\label{emiib}
\mathcal{L}_{MIIB}(\theta) = \mathcal{L}_{MS}(\theta) + \mathcal{L}_{MF}(\theta) + \mathcal{L}_{\text{synergy}}(\theta).
\end{align}

\subsection{Simplifying Optimization Objectives}

\wenlj{MIIB cannot be directly applied to multimodal representation learning because mutual information is generally intractable to compute. This difficulty is particularly pronounced when the encoders are deterministic: in this case, the relevant mutual information terms are not finite, a well-known obstacle for applying the information bottleneck principle to deterministic neural network classifiers~\cite{amjad2019learning}. Consequently, the mutual information objectives in Eq.~\eqref{emiib} become formally intractable, as they presuppose stochastic encoders capable of producing finite mutual information values.
The variational formulation of the information bottleneck~\cite{wang2022rethinking} typically requires the introduction of additional stochastic networks and has been observed to underperform in representation learning tasks. To address this limitation and enable MIIB for deterministic encoders, we exploit the relationship between mutual information and entropy to reformulate the objective as an entropy/alignment regularizer. This reformulation does not correspond to a rigorous stochastic information-compression bottleneck. Instead, it aligns more closely with the Deterministic Information Bottleneck (DIB) framework~\cite{DBLP:journals/neco/StrouseS17}, in which the entropy of the representation serves as a surrogate for the compression term.}

Consider that mutual information can be decomposed into entropy and conditional entropy as follows: $  I(A; B) = H(A) - H( A\mid B)  $. We can derive and simplify the mutual information term in MIIB. On the one hand, this approach converts the estimation of mutual information into the estimation of entropy and conditional entropy. On the other hand, it reduces computational overhead by merging similar terms.

\paragraph{\textbf{Simplifying the modality-specific loss }} 

The modality-specific loss is defined as
\begin{align}
\mathcal{L}_{MS}(\theta)&=\sum_{i=1}^{n}\Bigl\{\lambda_{\text{MS}} [I_\theta\left(Z'_i ; X'_i | X''\right)+I_\theta\left(Z''_i ; X''_i \mid X'\right)]\nonumber\\&- I_\theta\left(Z'_i ; X''\right) - I_\theta\left(Z''_i ; X'\right)\Bigr\}
\end{align}
Let's break it down step by step, focusing on the key terms inside the sum for each modality $  i  $.
First, recall that conditional mutual information $  I(A; B | C) = H(A | C) - H(A | B, C)  $. Applying this to the two terms:
\begin{align}
 &I_\theta\left(Z'_i ; X'_i | X''\right)+I_\theta\left(Z''_i ; X''_i \mid X'\right)
\nonumber\\
&=H(Z'_i | X'')-H(Z'_i |  X'_i,X'')\nonumber\\
&+H(Z''_i \mid X')-H(Z''_i \mid X''_i, X').
\end{align}
Here, $  Z'_i  $ is the representation from view $  X'_i  $ \wenlj{under the Markov assumption \((X'_i \to Z'_i, \text{target } X'')\)}, and $  Z''_i  $ from view $  X''_i  $ \wenlj{under the Markov assumption \((X''_i \to Z''_i, \text{target } X')\)}. 

\wenlj{In the deterministic encoder limit, the conditional distributions reduce to Dirac deltas, $  p(Z'_i \mid X'_i, X'') = \delta(Z'_i - f(X'_i))  $ and $  p(Z''_i \mid X''_i, X') = \delta(Z''_i - f(X''_i))  $. Consequently, the conditional differential entropies $  H(Z'_i \mid X'_i, X'')  $ and $  H(Z''_i \mid X''_i, X')  $ diverge to $  -\infty  $, causing the corresponding mutual information terms $  I_\theta(Z'_i; X'_i \mid X'')  $ and $  I_\theta(Z''_i; X''_i \mid X')  $ to diverge to $  +\infty  $.
Since the encoder is deterministic, the representation of each input view is fixed with no additional stochasticity. Following the approach of the Deterministic Information Bottleneck (DIB)~\cite{DBLP:journals/neco/StrouseS17}, these conditional entropy terms are therefore dropped from the objective, as they contribute no uncertainty once the input view is observed. In place of the original mutual information terms, we substitute the marginal conditional entropies $  H(Z'_i \mid X'')  $ and $  H(Z''_i \mid X')  $. This substitution can enforce information minimality (i.e., compression), yielding a tractable entropy-based objective.
So, simplifying, we get:
\begin{align}
 &I_\theta\left(Z'_i ; X'_i | X''\right)+I_\theta\left(Z''_i ; X''_i \mid X'\right)
\nonumber\\
& \rightarrow H(Z'_i | X'')+H(Z''_i \mid X'),
\end{align}

From a practical perspective, while the conditional distributions are nominally Dirac deltas (yielding infinite conditional entropy in the strict continuous case), real neural network implementations operate under finite floating-point precision, implicit stochasticity from Batch Normalization, weight decay, and quantization. These factors collectively act as an effective regularization scale $  \epsilon > 0  $, rendering the conditional entropy a large but finite negative constant that diverges only logarithmically with $  \epsilon  $~\cite{abdelaleem2025deep}. This practical regularization ensures that the resulting entropy-based objectives remain well-defined and tractable.
}

Next, let's handle the subtracted terms. Mutual information $  I(A; B) = H(A) - H(A | B)  $, so:
\begin{align}
&- I_\theta\left(Z'_i ; X''\right) - I_\theta\left(Z''_i ; X'\right)=\nonumber\\
&-H(Z'_i)+H(Z'_i | X'')-H(Z''_i)+H(Z''_i \mid X')
\end{align}
Now, putting it all together for the expression inside the loss 
\begin{align}
\mathcal{L}_{MS}(\theta)
=&\sum_{i=1}^{n}\Bigl\{-H(Z'_i)-H(Z''_i)\nonumber\\
&+(\lambda_{\text{MS}}+1)[H(Z'_i | X'')+H(Z''_i \mid X')]\Bigr\}.
\end{align}

\paragraph{\textbf{Simplifying the modality fusion loss}} 

The modality fusion loss function is defined as:
\begin{align}\label{MIB'}
\mathcal{L}_{MF}\left(\theta\right)= & \lambda_{\text{MF}} [I_\theta(Z'; X' \mid X'')+I_\theta(Z''; X'' \mid X') ]\nonumber\\
&-I_\theta(Z'; Z'') -\sum_{i=1}^n \left[ I_\theta(Z'_i; Z'') + I_\theta(Z''_i; Z')\right]\nonumber\\&- I_\theta(Z_s'; Z'') - I_\theta(Z_s''; Z').
\end{align}
Let's break it down step by step, following a similar approach to the previous derivation, by expanding terms using entropy definitions and assuming deterministic encoders.
First, recall that the conditional mutual information $  I(A; B | C) = H(A | C) - H(A | B, C)  $. Applying this to the two leading terms:
\begin{align}
 &I_\theta\left(Z'; X' | X''\right)+I_\theta\left(Z''; X'' \mid X'\right)
\nonumber\\
&=H(Z' | X'')-H(Z' | X',X'')\nonumber\\
&+H(Z'' \mid X')-H(Z'' \mid X'', X'),
\end{align}
Since the encoders and fusion are deterministic, the conditional distributions of $Z'$ given $(X',X'')$ and $Z''$ given $(X'',X')$ are degenerate. As in the modality-specific derivation, we drop the corresponding conditional differential entropy terms when constructing the surrogate objective, yielding:
\begin{align}
 &I_\theta\left(Z'; X' | X''\right)+I_\theta\left(Z''; X'' \mid X'\right)
\nonumber\\
&\rightarrow H(Z' | X'')+H(Z'' \mid X').
\end{align}
Now, let's handle the subtracted terms one by one. Mutual information $  I(A; B) = H(A) - H(A | B)  $, so we expand each using this form to highlight the conditional entropies:
First, $  - I_\theta(Z'; Z'') = -H(Z') + H(Z' | Z'')  $ (alternatively, it could be expressed as $  -H(Z'') + H(Z'' | Z')  $, but we'll use this version for consistency with cross terms).

Next, for the summed terms:
\begin{align}
&-\sum_{i=1}^n \left[ I_\theta\left(Z'_i ; Z''\right) + I_\theta\left(Z''_i ; Z'\right)\right]\nonumber\\
&=\sum_{i=1}^n \left[-H(Z'')+H(Z'' | Z'_i) - H(Z')+H(Z' \mid Z''_i)\right].
\end{align}
Similarly, for the final terms:
\begin{align}
&- I_\theta\left(Z_s'; Z''\right) - I_\theta\left(Z_s''; Z'\right)=\nonumber\\
&-H(Z_s')+H(Z_s' | Z'')-H(Z_s'')+H(Z_s'' \mid Z').
\end{align}
Hence, putting it all together, we obtain
\begin{align}
&\mathcal{L}_{MF}\left(\theta\right)= 
\lambda_{\text{MF}} \left[H(Z' | X'')+H(Z'' \mid X')\right]\nonumber\\
&-H(Z')+H(Z' | Z'')\nonumber\\
&+\sum_{i=1}^n \left[-H(Z'')+H(Z'' | Z'_i) - H(Z')+H(Z' \mid Z''_i)\right]\nonumber\\
&-H(Z_s')-H(Z_s'')+H(Z_s' | Z'')+H(Z_s'' \mid Z').
\end{align}

\paragraph{\textbf{Simplifying the synergistic representation loss}} 
The synergistic representation loss in multimodal fusion is defined as:
\begin{align}
\mathcal{L}_{\text{synergy}}\left(\theta\right)&=\lambda_{\text{ss}}[I_\theta\left(Z_{s}^{'}; X' | X''\right) + I_\theta\left(Z_{s}^{''}; X'' | X'\right)] \nonumber\\&+ \sum_{i=1}^n \left[ I_\theta(Z'_i; Z_s'') + I_\theta( Z''_i; Z_s')\right ].
\end{align}
This loss encourages conditional minimality of the synergistic representations and penalizes mutual information between individual modality representations and the opposite view's synergistic representation. Here, $  X' = \{X'_1, \dots, X'_n\}  $ and $  X'' = \{X''_1, \dots, X''_n\}  $ denote the multimodal inputs from two views, with $  n  $ modalities; $  Z'_i  $ and $  Z''_i  $ are the representations for modality $  i  $ from each view; and $  Z_s'  $ and $  Z_s''  $ are the fused synergistic representations from each view. Let's derive a simplified form step by step, assuming deterministic encoders (no added randomness), similar to the previous multi-view self-supervised loss derivation.
First, expand the conditional mutual information terms using the definition $  I(A; B | C) = H(A | C) - H(A | B, C)  $:
\begin{align}
&I_\theta\left(Z_{s}^{'}; X' | X''\right) + I_\theta\left(Z_{s}^{''}; X'' | X'\right) \nonumber\\
&= H(Z_s' | X'') - H(Z_s' | X', X'') \nonumber\\
&+ H(Z_s'' | X') - H(Z_s'' | X'', X').
\end{align}
Since the encoders and fusion are deterministic, the conditional distributions of $Z_s'$ given $(X',X'')$ and $Z_s''$ given $(X'',X')$ are degenerate. As in the modality-specific derivation, we drop the corresponding conditional differential entropy terms when constructing the surrogate objective, yielding:
\begin{align}
&I_\theta\left(Z_{s}^{'}; X' | X''\right) + I_\theta\left(Z_{s}^{''}; X'' | X'\right) \nonumber\\
& \rightarrow H(Z_s' | X'') + H(Z_s'' | X').
\end{align}

To enforce statistical independence in the disentangled representations, we require an effective estimator for the mutual information terms in the loss $  \sum_{i=1}^n \left[ I_\theta(Z'_i; Z_s'') + I_\theta(Z''_i; Z_s') \right]  $. Minimizing these mutual information quantities is equivalent to promoting orthogonality between the paired latent vectors in the representation space, which helps prevent entanglement while avoiding collapse modes where private representations align excessively with shared ones. Drawing from geometric constraints, we propose a cosine-similarity-based hinge loss that penalizes cosine similarities exceeding a threshold $\gamma$, thereby encouraging angular separation between unimodal and synergistic representations. This formulation offers computational efficiency over direct mutual information estimators and ensures robust disentanglement.
Specifically, the cosine disentanglement loss is defined as:
\begin{align}
\mathcal{L}_{\text{disentangle}} = \frac{1}{n} \sum_{i=1}^n \left[ \max\left(0, \frac{\langle \mathbf{z}'_i, \mathbf{z}_s'' \rangle}{\|\mathbf{z}'_i\|_2 \cdot \|\mathbf{z}_s''\|_2} - \gamma \right)\right] \nonumber\\
+ \frac{1}{n} \sum_{i=1}^n \left[\max\left(0, \frac{\langle \mathbf{z}''_i, \mathbf{z}_s' \rangle}{\|\mathbf{z}''_i\|_2 \cdot \|\mathbf{z}_s'\|_2} - \gamma \right) \right],
\label{eq:disentangle_loss}
\end{align}
where $\gamma$ controls the margin of the cosine-similarity constraint between unimodal and synergy representations.
This pairwise approach mirrors the independence objective. It is also differentiable almost everywhere and scalable for optimization in neural networks.
Combining all parts, the full loss becomes:
\begin{align}
\mathcal{L}_{\text{synergy}}\left(\theta\right) &= \lambda_{\text{ss}} \left[ H(Z_s' | X'') + H(Z_s'' | X') \right] \nonumber\\
&+ \mathcal{L}_{\text{disentangle}} .
\end{align}
\begin{algorithm}[t]
\caption{MIIB Training Algorithm}
\label{alg:our_training_batch}
\begin{algorithmic}[1]
    \REQUIRE Multimodal data augmentation $  \mathcal{T}^*  $,
independent unimodal encoders $  (f_i)_{i \in [1..n]}  $, where $  n  $ denotes the number of modalities,
fusion transformer $  g_f  $,
synergy transformer $  g_s  $.
    
    \FOR{sampling minibatch $ X = (X_1, \dots, X_n)$ }
      
        \STATE Draw $t', t'' \sim \mathcal{T}^*$ 
        \STATE $X', X'' \leftarrow t'(X), t''(X)$
        \FOR{$i \in [1, \dots, n]$}
            \STATE $e_i' \leftarrow f_i(X'_i)$
            \STATE $e_i'' \leftarrow f_i(X''_i)$
        \ENDFOR
        
        \STATE $Z' \leftarrow g_f(e'_1, \dots, e'_n)$
        \STATE $Z'' \leftarrow g_f(e''_1, \dots, e''_n)$
        
        \FOR{$i \in [1, \dots, n]$}
            \STATE $Z_i' \leftarrow g_f( e'_i)$
            \STATE $Z_i'' \leftarrow g_f(e''_i)$
        \ENDFOR
        \STATE $Z_s' \leftarrow g_s(e'_1, \dots, e'_n)$
        \STATE $Z_s'' \leftarrow g_s(e''_1, \dots, e''_n)$
        
        \STATE Update parameters $(f_i)_{i \in [1..n]}, g_f, g_s$ to minimize $\mathcal{\hat{L}}_{MIIB}$
    \ENDFOR
    \STATE \textbf{Return} $(f_i)_{i \in [1..n]}, g_f, g_s$
\end{algorithmic}
\end{algorithm}
This form emphasizes minimizing entropy in individual representations while conditioning on cross-view synergistic fusions for the synergistic representations themselves.

\paragraph{\textbf{Final Simplified Optimization Objectives}} 

We derive the variational upper bound for the conditional entropy in \wenlj{Appendix A}. The minimization of the conditional entropy is transformed into alignment between representations. For $  \mathcal{\hat{L}}_{MIIB}(\theta)  $, we obtain the final loss by merging the identical alignment and entropy terms.
\begin{align}
&\mathcal{\hat{L}}_{MIIB}(\theta) =\sum_{i=1}^n \left[-\lambda_{\text{single}}[H(Z'_i)+H(Z''_i)]\right]\nonumber\\
&-\lambda_{\text{fusion}}[H(Z')+H(Z'')]-\lambda_{\text{synergy}}[H(Z_s')+H(Z_s'')]\nonumber\\
&-\sum_{i=1}^n \left[\mathbb{E}_{p(Z'_i,Z'')}[Z_i^{'\top} Z''] +\mathbb{E}_{p(Z''_i,Z')}[Z_i^{''\top} Z']\right]\nonumber\\
&-\mathbb{E}_{p(Z',Z'')}[Z^{'\top} Z''] -\mathbb{E}_{p(Z_s',Z'')}[Z_s^{'\top} Z'']\nonumber\\
&-\mathbb{E}_{p(Z_s'',Z')}[Z_s^{''\top} Z']+ \mathcal{L}_{\text{disentangle}}.
\end{align}
We employ the Score-Based Entropy Estimation with the von Mises-Fisher Kernel from \cite{wen2024mveb} to estimate the entropy terms in $  \hat{\mathcal{L}}_{MIIB}(\theta) $, as described in detail in Appendix A. For the detailed MIIB Training Algorithm, please refer to Algorithm~\ref{alg:our_training_batch}.

\wenlj{\textbf{Relation to manifold-uniformity regularization}. Our entropy terms are related to alignment–uniformity\cite{pmlr-v119-wang20k}: maximizing the marginal hypersphere entropy $H(Z)$ is the uniformity objective (up to the bounded-support constant), and minimizing the conditional entropy $H(Z\mid X'')$ is the alignment objective. Thus MIIB's "bottleneck" is, by construction, a \textbf{structured multi-stream alignment–uniformity regularizer} that is principled, bounded, and consistent with the multi-view IB of \cite{federici2020learning} and our prior single-stream work \cite{wen2024mveb} rather than a literal channel-capacity constraint. }

\section{Experiments}
  We designed different sets of experiments to evaluate the effectiveness of MIIB, encompassing both synthetic benchmarks and multiple large-scale real-world datasets. First, to evaluate the model's capacity to capture three core multimodal interactions (i.e., uniqueness, redundancy, and synergy), we constructed a controlled environment based on the Trifeature dataset~\cite{hermann2020shapes}. We followed the design in CoMM~\cite{dufumier2024align} to generate synthetic data and assessed performance through specifically designed tasks. Second, we conducted experiments on several widely used multimodal benchmark datasets spanning real-world scenarios in diverse domains (e.g., healthcare, robotics, etc.), aiming to evaluate the model's generalizability under different modality combinations and demonstrate its capability to learn effective multimodal representations for complex tasks. Detailed experimental settings are provided in Appendix B. Unless otherwise specified, we adopted the linear probing protocol for evaluation. Specifically, we freeze the pre-trained feature extractors and train a corresponding linear classifier or regressor on top of the learned representations, depending on the specific downstream task.

\subsection{Controlled Experiments on Synthetic Trifeature}

Following the experimental protocol of CoMM~\cite{dufumier2024align}, we construct a synthetic bimodal dataset based on Trifeature to establish a controlled environment, evaluating MIIB's capacity to capture uniqueness, redundancy, and synergy through two independent experimental settings.
\begin{table}[h]
    \centering
    \caption{Linear probing accuracy (in \%) of redundancy (shape), uniqueness (texture) and synergy (color and texture) on Trifeature dataset. $^{\dagger}$ denotes results are from \cite{dufumier2024align}.}
    \label{tab:trifeature_results}
    \begin{tabular}{l c c c}
        \toprule
        Model & Redundancy$\uparrow$ & Uniqueness$\uparrow$ & Synergy$\uparrow$ \\
        \midrule
        Cross$^{\dagger}$~\cite{radford2021learning}  & $\mathbf{100.0}$ & $11.6$ & $50.0$ \\
        Cross+Self$^{\dagger}$~\cite{yuan2021multimodal}  & $99.7$ & $86.9$ & $50.0$ \\
        FactorCL$^{\dagger}$~\cite{liang2023factorized}  & $99.8$ & $62.5$ & $46.5$ \\
        CoMM~\cite{dufumier2024align}  & $99.9_{\pm0.06}$ & $86.8_{\pm2.99}$ & $71.4_{\pm3.47}$ \\
        MIIB (ours) & $99.9_{\pm0.12}$ & $\mathbf{89.8}_{\pm3.23}$ & $\mathbf{79.1}_{\pm2.97}$ \\
        \bottomrule
    \end{tabular}
\end{table}

For the assessment of uniqueness and redundancy, we define shape as the shared (redundant) feature and texture as the modality-specific (unique) feature. We generate the dataset by selecting image pairs that share an identical shape but possess independent textures. We then design two linear probing sub-tasks: (1) identifying the shared shape class (redundancy task); and (2) predicting the modality-specific unique texture class (uniqueness task). The chance-level accuracy for both tasks is $10\%$.

For the evaluation of synergy, we artificially construct a strong cross-modal correlation within the training set by defining a mapping $\mathcal{M}$ between the texture of Modality 1 ($X_1$) and the color of Modality 2 ($X_2$) (e.g., stripes correspond to red). The model is trained exclusively on image pairs that satisfy this mapping, with the downstream task defined as determining whether a given image pair complies with the mapping rule, formulated as $Y=1 \iff (\text{texture}(X_1), \text{color}(X_2)) \in \mathcal{M}$. The chance-level accuracy for this task is $50\%$.

The quantitative results are summarized in Table~\ref{tab:trifeature_results}. The ``Cross'' model with InfoNCE-based cross-modality constraints~\cite{radford2021learning} achieves the best performance in capturing redundant information but struggles with uniqueness and synergy. FactorCL~\cite{liang2023factorized} and self-supervised constraints on each encoder (``Cross+Self'')~\cite{yuan2021multimodal} enhance uniqueness modeling but remain limited in synergy. CoMM~\cite{dufumier2024align} serves as a comprehensive baseline by addressing all three task-related interactions. However, the synergy accuracy of $71.4\%$ suggests that there remains substantial room for more effectively capturing synergistic information. In contrast, MIIB demonstrates superior performance in encoding both uniqueness and synergy, yielding improvements of $3.0$ and $7.7$ percentage points over CoMM, respectively. This improvement can be attributed to the proposed synergy constraint, which further enhances the extraction of collaborative multimodal information, thereby better capturing complex cross-modal interactions.

\subsection{Experiments on MultiBench}
Following~\cite{liang2023factorized,dufumier2024align}, we evaluate on a subset of real-world datasets from MultiBench~\cite{liang2021multibench} spanning diverse domains: (1) the MIMIC dataset~\cite{johnson2016mimic} for healthcare, including static tabular data and ICU time-series; (2) MOSI~\cite{zadeh2016multimodal}, UR-FUNNY~\cite{hasan2019ur}, and MUSTARD~\cite{castro2019towards} for affective computing and multimedia analysis, including visual, audio, and textual modalities; and (3) the Vision \& Touch dataset~\cite{lee2020making} for robotics manipulation, including image, force, and proprioception data. Further details regarding the datasets are provided in Appendix C.
\paragraph{Performance on Bimodal Benchmarks} For a fair comparison, we utilize identical encoders and modality configurations as CoMM. \wenlj{We evaluate MIIB under two settings: (1) self-supervised pretraining followed by linear probing and (2) supervised end-to-end fine-tuning after pretraining. Under the linear probing protocol,} we compare MIIB against several strong baselines, including Cross, Cross+Self, FactorCL, and CoMM. \wenlj{In the fine-tuning setting, we focus on CoMM, the most closely related method, and conduct a controlled comparison using the same fine-tuning protocol.}
\begin{table*}[h]
    \centering
    \caption{Linear probing MSE($\times 10^{-4}$) for regression task and top-1 accuracy (in \%) for classification tasks on Multibench. $^{\dagger}$ denotes results are from~\cite{dufumier2024align}. $^*$ denotes average is only selected from the results of classification tasks. \wenlj{Colored rows indicate fine-tuning results.}}
    \label{tab:multibench_results}
    \setlength{\tabcolsep}{10pt}
    
    \begin{tabular}{lcccccc}
        \toprule
        \multicolumn{1}{c}{\multirow{2}{*}{Model}} & Regression & \multicolumn{5}{c}{Classification} \\
        \cmidrule(lr){2-2} \cmidrule(lr){3-7}
              & V\&T EE$\downarrow$ 
              & MIMIC$\uparrow$ 
              & MOSI$\uparrow$ 
              & UR-FUNNY$\uparrow$ 
              & MUSTARD$\uparrow$ 
              & Average$^*$ $\uparrow$ \\
        \midrule
        Cross$^{\dagger}$~\cite{radford2021learning} & $33.09_{\pm 3.67}$ & $66.7_{\pm 0.1}$ & $47.8_{\pm 1.8}$ & $50.1_{\pm 1.9}$ & $53.5_{\pm 2.9}$ & $54.52$ \\
        Cross+Self$^{\dagger}$~\cite{yuan2021multimodal} & $7.56_{\pm 0.31}$ & $65.49_{\pm 0.0}$ & $49.0_{\pm 1.1}$ & $59.9_{\pm 0.9}$ & $53.9_{\pm 4.0}$ & $57.07$ \\
        FactorCL$^{\dagger}$~\cite{liang2023factorized} & $10.82_{\pm 0.56}$ & $67.3_{\pm 0.0}$ & $51.2_{\pm 1.6}$ & $60.5_{\pm 0.8}$ & $55.80_{\pm 0.9}$ & $58.7$ \\
        CoMM~\cite{dufumier2024align} & $7.96_{\pm 2.13}$ & $66.4_{\pm 0.4}$ & $63.7_{\pm 2.5}$ & $63.3_{\pm 0.5}$ & $64.4_{\pm 1.1}$ & $64.45$ \\
        MIIB (ours) & $\mathbf{5.74}_{\pm 0.69}$ & $\mathbf{68.1}_{\pm 0.3}$ & $\mathbf{67.4}_{\pm 1.8}$ & $\mathbf{65.4}_{\pm 0.9}$ & $\mathbf{66.0}_{\pm 2.0}$ & $\mathbf{66.73}$ \\
        \midrule
        \rowcolor{cyan!10} CoMM~\cite{dufumier2024align} & $1.61_{\pm 0.16}$ & $67.57_{\pm 0.65}$ & $71.84_{\pm 2.7}$ & $64.99_{\pm 0.96}$ & $68.25_{\pm 1.69}$ & $68.16$ \\
        \rowcolor{cyan!10} MIIB(fine-tuned) & $\mathbf{1.47}_{\pm 0.06}$ & $\mathbf{69.0}_{\pm 0.57}$ & $\mathbf{73.24}_{\pm 1.27}$ & $\mathbf{66.24}_{\pm 0.18}$ & $\mathbf{69.73}_{\pm 2.30}$ & $\mathbf{69.55}$ \\
        \bottomrule
    \end{tabular}
\end{table*}

As presented in Table~\ref{tab:multibench_results}, \wenlj{under the linear probing setting,} MIIB achieves the best performance across all evaluated datasets. In classification tasks, MIIB surpasses the CoMM baseline in average accuracy ($66.73\%$ vs. $64.45\%$). Specifically, we observe substantial improvements of $3.7$, $2.1$, $1.6$, and $1.7$ percentage points on the MOSI, UR-FUNNY, MUSTARD, and MIMIC datasets. For the regression task, MIIB achieves an MSE of $5.74 (\times 10^{-4})$, outperforming the previous best result of $7.56$. \wenlj{In the supervised full-parameter fine-tuning setting, MIIB consistently achieves better performance across all evaluated benchmarks, improving the average score over CoMM from 68.16 to 69.55.}

Overall, these results demonstrate the effectiveness of MIIB and highlight its robustness across diverse real-world benchmarks under bimodal settings.
\paragraph{Performance on Trimodal Benchmarks} To assess the generalization of MIIB to more than two modalities, we conduct experiments on two trimodal datasets from MultiBench: Vision\&Touch (vision, force, and proprioception modalities for contact prediction) and UR-FUNNY (vision, text, and audio for humor detection). 
To better evaluate the information gain from the third modality, we include bimodal results for contrast, using visual and proprioceptive modalities for Vision\&Touch and textual and visual modalities for UR-FUNNY. 
In the bimodal setting, we compare against Cross, Cross+Self, and CoMM; in the trimodal setting, we adopt CMC~\cite{tian2020cmc} and CoMM as baselines.

Results are summarized in Table~\ref{tab:comparison_sota}. First, incorporating the additional modality yields consistent gains for MIIB: accuracy gains of $5.8$ percentage points on Vision\&Touch and $0.8$ percentage points on UR-FUNNY. 
Second, MIIB achieves comparable or superior performance to all baselines in trimodal evaluations. 
Notably, on UR-FUNNY, it surpasses the leading baseline CoMM by $1.4$ percentage points; on Vision\&Touch, \wenlj{MIIB matches the state-of-the-art performance of CoMM.} These results indicate that our model can be extended beyond bimodal settings, effectively utilizing additional modalities to further enhance performance.
\begin{table}[t]
    \centering
    \caption{Linear probing top-1 accuracy (in \%) for classification tasks on Vision\&Touch and UR-FUNNY. $^{\dagger}$ denotes results are from \cite{dufumier2024align}}
    \label{tab:comparison_sota}
    
    \renewcommand{\arraystretch}{1.1}
    \setlength{\tabcolsep}{6pt}
    
    \begin{tabular}{lccc}
        \toprule
        \textit{Model} & \textit{\#Mod.} & \textit{V\&T CP}$\uparrow$ & \textit{UR-FUNNY}$\uparrow$ \\
        \midrule
        Cross~\cite{radford2021learning} & 2 & $86.3_{\pm 0.25}$ & $50.1^{\dagger}$ \\
        Cross+Self~\cite{yuan2021multimodal} & 2 & $87.6_{\pm 0.26}$ & $59.9^{\dagger}$ \\
        CoMM~\cite{dufumier2024align} & 2 & $87.0_{\pm 1.58}$ & $63.3_{\pm 0.51}$ \\
        MIIB (ours) & 2 & $\mathbf{88.4}_{\pm 0.32}$ & $\mathbf{65.4}_{\pm 0.90}$ \\
        \midrule
        CMC$^{\dagger}$~\cite{tian2020cmc}  & 3 & $94.1$ & $59.2$ \\
        CoMM~\cite{dufumier2024align} & 3 & $94.1_{\pm 0.17}$ & $64.8_{\pm 1.13}$ \\
        MIIB (ours) & 3 & $\mathbf{94.2}_{\pm 0.11}$ & $\mathbf{66.2}_{\pm 0.85}$ \\
        \bottomrule
    \end{tabular}
\end{table}

\subsection{Experiments on Multimodal IMDb}
MM-IMDb~\cite{arevalo2017gated} is a real-world multi-label movie genre prediction dataset consisting of poster images and plot summaries. 
The task is challenging due to the highly imbalanced class distribution and the large semantic gap between the visual and textual modalities. 
Consistent with prior work~\cite{liang2021multibench}, the visual (posters) and textual (plot summaries) modalities provide complementary cues—capturing aesthetic style and narrative content, respectively. Consequently, accurate genre prediction requires synergizing information from both modalities. \wenlj{We evaluate the proposed MIIB under two experimental settings: self-supervised linear probing and supervised fine-tuning. In the self-supervised setting,} we compare against unimodal baselines (SimCLR~\cite{chen2020simple}, CLIP (Visual), and CLIP (Text)~\cite{radford2021learning}) and multimodal baselines (SLIP~\cite{mu2022slip}, CLIP (Visual + Text), and CoMM). 
For a fair comparison, MIIB and CoMM are built on the same pretrained CLIP backbone. \wenlj{In the supervised fine-tuning setting, we compare MIIB against representative fully supervised baselines such as MFAS~\cite{perez2019mfas} and ReFNet~\cite{sankaran2022refining}. In addition, we compare MIIB with CoMM under an identical full fine-tuning protocol.}

As shown in Table~\ref{tab:mm_imdb_results}, MIIB achieves the best performance across all evaluation metrics. 
Compared to the strong multimodal baseline CoMM, we obtain improvements of $1.51$ percentage points in Weighted F1 and $2.41$ percentage points in Macro F1.
We also observe that CLIP with its original pre-trained weights outperforms CLIP further trained on unlabeled MM-IMDb data ($58.9\%$ vs. $54.49\%$ in Weighted F1). 
This observation suggests that relying solely on redundancy-focused alignment may be insufficient for movie genre prediction, which often benefits from complementary integration of visual and textual modalities~\cite{dufumier2024align}. In contrast, the superior performance of MIIB highlights its effectiveness on this imbalanced and semantically heterogeneous multi-label multimodal task. \wenlj{In the fine-tuning setting, MIIB consistently outperforms both existing fully supervised baselines and fine-tuned CoMM.}

\begin{table}[t]
    \centering
    \caption{Linear probing F1-score (weighted and macro) (in \%) for MM-IMDb. $\triangle$ indicates further training on unlabeled data. $\dagger$ denotes results are from \cite{dufumier2024align}. For fair comparison, both CoMM and MIIB are implemented using the same pre-trained CLIP backbone. \wenlj{Colored rows indicate fine-tuning results.}}
    \label{tab:mm_imdb_results}

    \begin{tabular}{lccc}
        \toprule
        \textit{Model} & \textit{Modalities} & \textit{weighted-f1}$\uparrow$ & \textit{macro-f1}$\uparrow$ \\
        \midrule
        SimCLR$^{\dagger \triangle}$~\cite{chen2020simple} & V & $40.35_{\pm 0.23}$ & $27.99_{\pm 0.33}$ \\
        CLIP$^{\dagger}$~\cite{radford2021learning} & V & $51.5$ & $40.8$ \\
        CLIP$^{\dagger}$~\cite{radford2021learning} & L & $51.0$ & $43.0$ \\
        CLIP$^{\dagger}$~\cite{radford2021learning} & V+L & $58.9$ & $50.9$ \\
        SLIP$^{\dagger \triangle}$~\cite{mu2022slip} & V+L & $56.54_{\pm 0.19}$ & $47.35_{\pm 0.27}$ \\
        CLIP$^{\dagger \triangle}$~\cite{radford2021learning} & V+L & $54.49_{\pm 0.19}$ & $44.94_{\pm 0.30}$ \\
        CoMM~\cite{dufumier2024align} & V+L & $61.29_{\pm 0.73}$ & $53.79_{\pm 0.22}$ \\
        MIIB (ours) & V+L & $\mathbf{62.80}_{\pm 0.24}$ & $\mathbf{56.20}_{\pm 0.16}$ \\
        \midrule
        \rowcolor{cyan!10} MFAS$^{\dagger}$~\cite{perez2019mfas} & V+L & 62.50 & 55.6 \\
        \rowcolor{cyan!10}ReFNet$^{\dagger}$ ~\cite{sankaran2022refining} & V+L & - & 56.7 \\
        \rowcolor{cyan!10} CoMM~\cite{dufumier2024align} & V+L & $63.92_{\pm 0.41}$ & $57.90_{\pm 0.39}$ \\
        \rowcolor{cyan!10} MIIB (ours) & V+L & $\mathbf{65.14}_{\pm 0.11}$ & $\mathbf{59.32}_{\pm 0.33}$ \\
        \bottomrule
    \end{tabular}
    \par
    \vspace{-0.5em}
\end{table}

    

    
\wenlj{\subsection{Few-shot Transfer Performance on the MM-IMDb Dataset}}

\wenlj{To further evaluate the transferability and sample efficiency of the learned representations, we conduct few-shot experiments on the MM-IMDb dataset using only 1\% and 10\% of the labeled training data, together with a full fine-tuning setting for comparison. We restrict this few-shot evaluation to MM-IMDb, which contains 25,959 movies in total and is substantially larger than the affective-computing benchmarks considered here, such as MUSTARD (690 utterances) and MOSI (2,199 utterances). As shown in Table~\ref{tab:mmimdb_fewshot_vertical}, MIIB consistently outperforms CoMM across all supervision levels in terms of both Weighted-F1 and Macro-F1. Notably, under only 1\% labeled data, MIIB outperforms CoMM by 2.54 in Weighted-F1 and 3.14  in Macro-F1, achieving scores of 63.22 and 56.80, respectively. The consistent improvements under the 10\% and full-data settings further indicate that MIIB learns transferable multimodal representations that remain effective across different levels of supervision.}

\begin{table}[h]
    \centering
    \caption{\wenlj{Performance comparison on the MM-IMDb dataset under different evaluation protocols. Results are reported in Macro-F1 and Weighted-F1 along with standard deviations. ``Linear'' denotes the best self-supervised linear probing results. }}
    \label{tab:mmimdb_fewshot_vertical}
    \setlength{\tabcolsep}{10pt} 
    \begin{tabular}{clcc}
        \toprule
        \textit{Setting} & \textit{Model} & \textit{weighted-f1}$\uparrow$ & \textit{macro-f1}$\uparrow$ \\
        \midrule
        \multirow{2}{*}{Linear} & CoMM~\cite{dufumier2024align} & $61.29_{\pm 0.73}$ & $53.79_{\pm 0.22}$ \\
        & MIIB (ours) & $\mathbf{62.80}_{\pm 0.24}$ & $\mathbf{56.20}_{\pm 0.16}$ \\
        \midrule
        \multirow{2}{*}{1\%} & CoMM~\cite{dufumier2024align} & $60.68_{\pm 0.31}$ & $53.66_{\pm 0.40}$ \\
        & MIIB (ours) & $\mathbf{63.22}_{\pm 0.31}$ & $\mathbf{56.80}_{\pm 0.47}$ \\
        \midrule
        \multirow{2}{*}{10\%} & CoMM~\cite{dufumier2024align} & $61.60_{\pm 0.41}$ & $54.48_{\pm 0.72}$ \\
        & MIIB (ours) & $\mathbf{63.60}_{\pm 0.26}$ & $\mathbf{57.00}_{\pm 0.47}$ \\
        \midrule
        \multirow{2}{*}{Full} & CoMM~\cite{dufumier2024align} & $63.92_{\pm 0.41}$ & $57.90_{\pm 0.39}$ \\
        & MIIB (ours) & $\mathbf{65.14}_{\pm 0.11}$ & $\mathbf{59.32}_{\pm 0.33}$ \\
        \bottomrule
    \end{tabular}
\end{table}

\begin{table}[t]
\centering
\caption{
\wenlj{InterSHAP analysis over 15 random seeds on two benchmark datasets.
Results are reported as mean $\pm$ standard deviation together with paired $t$-tests.}
}
\label{tab:intershap_stats}
\small
\setlength{\tabcolsep}{12pt}
\begin{tabular}{lcc}
\toprule
Model Setting & UR-FUNNY & MIMIC \\
\midrule
w/o Synergy & $5.78_{\pm 2.05}$ & $1.34_{\pm 0.80}$  \\
w/ Synergy  & $\mathbf{7.32}_{\pm 1.98}$ & $\mathbf{2.54}_{\pm 1.26}$ \\
\midrule
Improvement & $\mathbf{+1.54}$ & $\mathbf{+1.20}$ \\
t-test & $2.70$ & $3.61$ \\
$p$-value   & $0.018$ & $0.003$ \\
\bottomrule
\end{tabular}
\end{table}

\subsection{Quantitative Analysis of Cross-Modal Interactions via InterSHAP}
To rigorously quantify the extent to which our proposed synergy constraint facilitates effective cross-modal interactions, we employ the InterSHAP metric~\cite{wenderoth2025measuring}. Unlike traditional performance-based metrics (e.g., accuracy or F1 score) that only reflect the final outcome, InterSHAP is designed to explicitly disentangle and measure the contribution of \textit{interactions} versus \textit{unimodal} features to the model's decision-making process. 
InterSHAP quantifies the proportion of a model's behavior attributable to cross-modal interactions, leveraging the Shapley Interaction Index (SII) to decompose the prediction $f(x)$ into individual modality contributions (main effects) and cross-modal interaction terms. Let $M$ be the number of modalities and $\phi_{ij}$ denote the SII value between modalities $i$ and $j$ (aggregated globally across the dataset or locally per sample). The InterSHAP score is defined as the ratio of the aggregated interaction effects to the total model behavior:
\begin{equation}
    \text{InterSHAP} = \frac{\sum_{\substack{i,j=1 \\ i \neq j}}^{M} |\phi_{ij}|}{\sum_{i,j=1}^{M} |\phi_{ij}|},
\end{equation}
where the numerator sums the absolute interaction effects ($i \neq j$), and the denominator represents the total absolute attributions, including unimodal main effects ($i = j$). A higher InterSHAP value indicates greater reliance on cross-modal interaction effects rather than unimodal main effects. This metric supports any M, provides both local (sample-level) and global explanations, and is performance-agnostic.


\wenlj{As shown in Table~\ref{tab:intershap_stats}, we extend the InterSHAP analysis to 15 random seeds and conduct paired $t$-tests to assess statistical significance. The proposed synergy objective consistently increases InterSHAP values on both datasets, improving the interaction ratio from 5.78\% to 7.32\% on UR-FUNNY and from 1.34\% to 2.54\% on MIMIC. Moreover, the corresponding improvements are statistically significant ($p<0.05$), suggesting that the observed interaction gains are robust across different random initializations. Since InterSHAP is designed to quantify the relative contribution of cross-modal interaction effects, these results provide strong empirical evidence that the proposed synergy objective effectively encourages richer cross-modal complementary interactions.} \wenlj{InterSHAP estimation variance is governed primarily by test-set size: on the two datasets above (UR-FUNNY, $N=1058$; MIMIC, $N=3262$), the per-seed variance is small enough for the synergy-induced gain to be clearly resolved. We additionally applied the same 15-seed analysis to MOSI ($N=686$) and MUSTARD ($N=138$); at this much smaller scale, the sampling variance of the Shapley interaction estimate itself becomes too large for any method comparison to be statistically reliable, independent of whether the underlying synergy effect is real. We therefore restrict our significance claim to the datasets on which InterSHAP can be reliably measured, and report the MOSI/MUSTARD results in full in Appendix B rather than omitting them.}

\subsection{Ablation Studies}
To rigorously evaluate the effectiveness of MIIB, we conduct ablation studies on two challenging real-world benchmarks: UR-FUNNY and MUSTARD. We select these datasets because, compared to synthetic benchmarks (e.g., Trifeature), affective computing tasks such as humor and sarcasm detection typically require more complex cross-modal interactions (e.g., semantic incongruity). Recent information-theoretic analysis~\cite{liang2023quantifying} further shows that UR-FUNNY and MUSTARD exhibit higher levels of synergy under their proposed metric than many commonly used multimodal benchmarks.
Specifically, our ablation study consists of two parts: loss function ablation and parameter sensitivity analysis.
\begin{table}[h]
    \centering
    \caption{Linear probing accuracy (in \%) on \textsc{UR-FUNNY} and \textsc{MUSTARD} datasets under different combinations of loss functions. $\lambda_1$, $\lambda_2$, and $\lambda_3$ denote the weights for $\mathcal{L}_{MF}$ , $\mathcal{L}_{MS}$, and $\mathcal{L}_{synergy}$}
    \label{tab:ablation_comparison}
    
    \setlength{\tabcolsep}{8pt}
    
    \begin{tabular}{ccccc}
        \toprule
        \multicolumn{3}{c}{\textit{Loss Weights}} & \multirow{2}{*}{\textsc{UR-FUNNY}\,$\uparrow$} & \multirow{2}{*}{\textsc{MUSTARD}\,$\uparrow$} \\
        \cmidrule(r){1-3} 
        $\lambda_1$ & $\lambda_2$ & $\lambda_3$ &  \\
        \midrule
        1 & 0 & 0 & $63.5_{\pm 1.07}$ & $62.4_{\pm 3.22}$ \\
        1 & 1 & 0 & $63.8_{\pm 0.90}$ & $65.2_{\pm 1.24}$\\
        1 & 0 & 1 & $63.9_{\pm 0.71}$ & $62.8_{\pm 1.61}$\\
        1 & 1 & 1 & $\mathbf{65.4}_{\pm 0.89}$ & $\mathbf{66.0}_{\pm 2.0}$ \\
        \bottomrule
    \end{tabular}
    \vspace{-0.5em}
\end{table}

\begin{figure}[h] 
    \centering
    
    \includegraphics[width=0.48\textwidth]{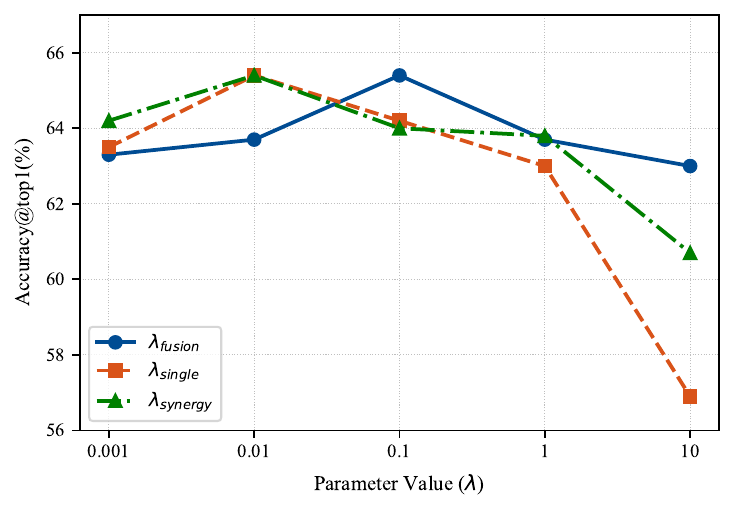}

    \vspace{-0.1in} 
    
    \caption{Parameter sensitivity analysis on the UR-FUNNY dataset with respect to the fusion penalty ($\lambda_{\text{fusion}}$), unimodality penalty ($\lambda_{\text{single}}$), and the synergy penalty ($\lambda_{\text{synergy}}$). }
    \label{fig:sensitivity}
    \vspace{-0.0in}
\end{figure}

    

    

\begin{figure}[h] 
    \centering
    
    \includegraphics[width=0.48\textwidth]{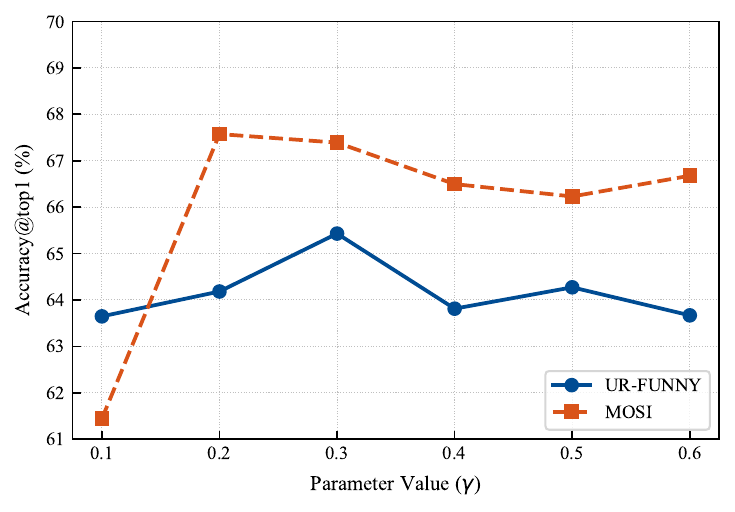}

    \vspace{-0.1in} 
    
    \caption{\wenlj{Parameter sensitivity analysis of the hyperparameter $\gamma$ on the UR-FUNNY and MOSI datasets. }}
    \label{fig:gamma}
    \vspace{-0in}
\end{figure}
\paragraph{Loss Function} \wenlj{Table ~\ref{tab:ablation_comparison} presents the ablation results for different loss components in MIIB. Using only $\lambda_1$ (i.e., only $\mathcal{L}_{MF}$) yields the lowest performance on both datasets. Enabling the unimodal constraint ($\lambda_2=1$) consistently improves the results, with a particularly notable gain on MUSTARD (62.4\% $\rightarrow$ 65.2\%). Similarly, enabling the synergy objective ($\lambda_3=1$) also leads to performance improvements over the baseline, demonstrating its effectiveness in modeling cross-modal interactions. When both objectives are jointly incorporated, MIIB achieves the best results of 65.4\% on UR-FUNNY and 66.0\% on MUSTARD, indicating that the unimodal constraint and the synergy objective provide complementary benefits for learning robust multimodal representations.}

\begin{table*}[t]
    \centering
    \caption{Comparative analysis of computational complexity and training efficiency. We report FLOPs, MACs, model size (\#Params), and total training time. All measurements are conducted on a single NVIDIA 4090 GPU}
    \label{tab:complexity_efficiency}
    \begin{tabular*}{\textwidth}{l @{\extracolsep{\fill}} lcccc}
    \toprule
    Dataset & Model & FLOPs & MACs & \#Params & Total Training Time \\
    \midrule
    \multirow{2}{*}{UR-FUNNY} & CoMM & $0.15\,\mathrm{G}$ & $0.08\,\mathrm{G}$ & $7.11\,\mathrm{M}$ & $0.48\,\mathrm{h}$  \\
                              & MIIB & $0.18\,\mathrm{G}$ & $0.09\,\mathrm{G}$ & $7.13\,\mathrm{M}$ & $0.54\,\mathrm{h}$ \\
    \midrule
    \multirow{2}{*}{MMIMDb-clip} & CoMM & $36.27\,\mathrm{G}$ & $18.14\,\mathrm{G}$ & $236.5\,\mathrm{M}$ & $4.11\,\mathrm{h}$  \\
                                 & MIIB & $38.08\,\mathrm{G}$ & $19.04\,\mathrm{G}$ & $243.6\,\mathrm{M}$ & $4.31\,\mathrm{h}$ \\
    \bottomrule
    \end{tabular*}
\end{table*}
\begin{figure*}[th] 
    \centering
    \includegraphics[width=0.75\textwidth]{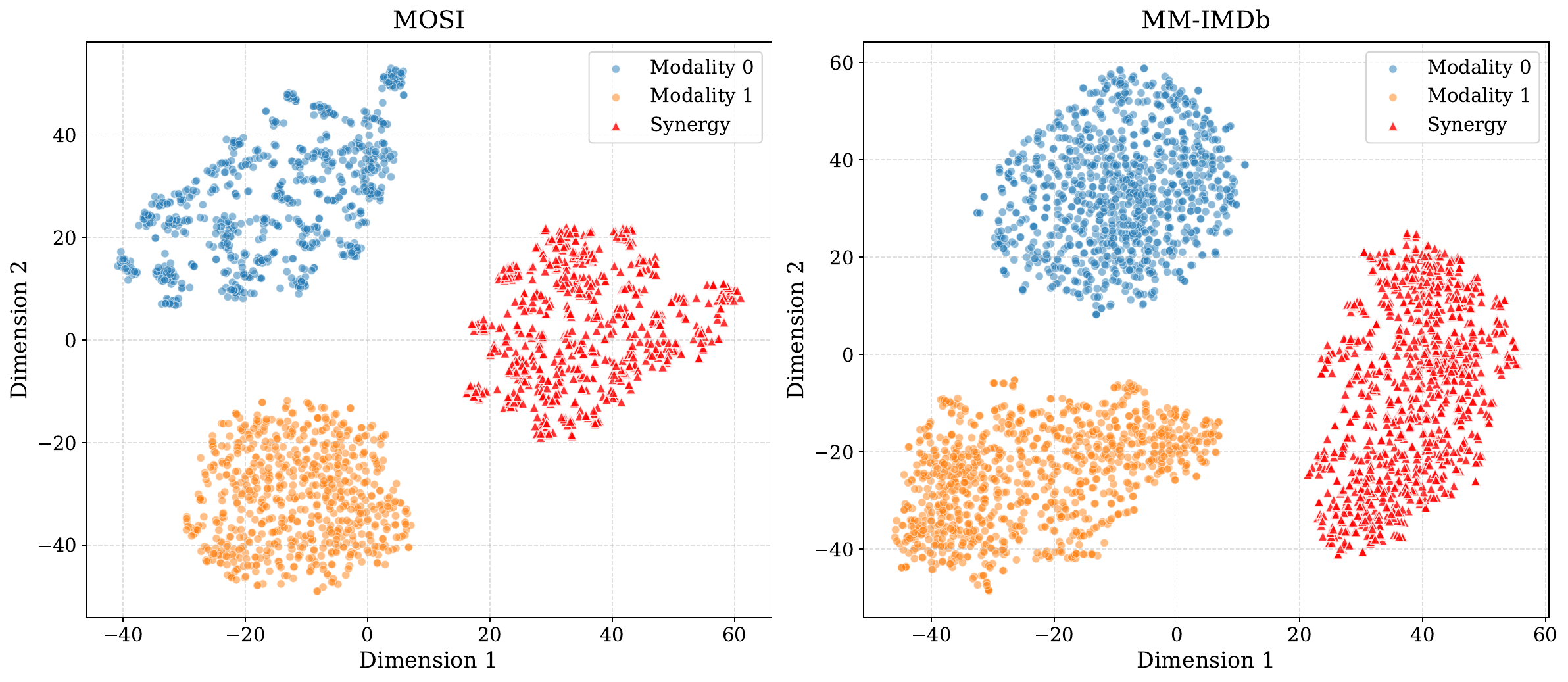} 
    \vspace{-0.1in} 
    \caption{t-SNE visualization of unimodal and synergy representations on the MOSI (left) and MM-IMDb (right) datasets. The figure illustrates the distribution of unimodal representations (Modality 0, Modality 1) and the corresponding synergy representations in the learned embedding space. }
    \label{fig:disentanglement}
    
    \vspace{-0.1in}
\end{figure*}
\paragraph{Parameter sensitivity analysis} 
\wenlj{We analyze the sensitivity of the key hyperparameters in MIIB. 
Figure~\ref{fig:sensitivity} presents the results on the UR-FUNNY dataset for the loss weights $\lambda_{\text{fusion}}$, $\lambda_{\text{single}}$, and $\lambda_{\text{synergy}}$, which correspond to the entropy regularization coefficients for multimodal alignment, unimodal alignment, and synergistic representations, respectively. All three parameters exhibit non-monotonic trends, indicating that moderate values generally yield better performance than extreme settings.}
In particular, the unimodality penalty $\lambda_{\text{single}}$ is the most sensitive: overly large values lead to a pronounced performance drop, indicating that excessively strong entropy regularization on unimodal representations can overly flatten the feature distribution and reduce discriminative power.
In contrast, $\lambda_{synergy}$ yields relatively stable performance across a broad range of values, suggesting robustness to its scaling.
The fusion penalty $\lambda_{\text{fusion}}$ reaches its best performance around $\lambda_{\text{fusion}}=0.1$, with performance degrading when it is set too small or too large.

\wenlj{We further investigate the effect of the margin parameter $\gamma$, which controls the cosine-similarity constraint between unimodal and synergy representations. As shown in Fig.~\ref{fig:gamma}, both UR-FUNNY and MOSI exhibit non-monotonic trends with respect to $\gamma$, with the best performance obtained at moderate values ($\gamma\approx0.2$--$0.3$). Performance decreases when $\gamma$ is set either too small or too large, indicating that downstream performance is sensitive to the choice of the margin parameter.}

\subsection{Computational Complexity and Runtime Analysis}
As shown in Table~\ref{tab:complexity_efficiency}, MIIB maintains a computational complexity comparable to the CoMM baseline. The slight increase in complexity is primarily attributed to the synergy transformer, which adopts the same architectural design as the fusion transformer. Despite this structural addition, the resulting overhead remains modest relative to the CoMM baseline. For instance, on the large-scale MM-IMDb dataset, the parameter count and FLOPs increase by roughly 3--5\% (e.g., params from 236.5M to 243.6M), with a corresponding increase in training time of roughly 12 minutes. We consider this slight computational cost to be a justifiable trade-off given the significant improvements in performance.

\subsection{Visualization of Unimodal and Synergy Representational Disentanglement}
To examine whether the proposed disentanglement loss encourages separation among different representations, we visualize unimodal and synergy representations using t-SNE on the MOSI and MM-IMDb datasets, as shown in Figure \ref{fig:disentanglement}. 
As observed, unimodal representations from different modalities and the synergy representation occupy clearly separated regions in the embedding space on both datasets. This separation reflects the explicit constraint imposed by Eq. \eqref{eq:disentangle_loss}, which discourages excessive similarity between unimodal and synergy representations and promotes their separation. These visualizations provide qualitative evidence that the proposed objective encourages disentanglement between unimodal and synergy representations.

\section{Conclusion}
\wenlj{In this work, we formalize MIIB from an information-theoretic perspective. We show that modality-specific learning and inter-modal fusion dependencies fundamentally limit existing multimodal methods. By employing PID to encode redundancy, uniqueness, and synergy, MIIB introduces a dual-path Transformer architecture with entropy-based objectives that produce minimal yet sufficient multimodal representations, consistently outperforming strong baselines across multiple benchmarks.

We acknowledge that MIIB introduces a modest computational overhead: on MM-IMDb, the dedicated Synergy Transformer increases parameters and FLOPs by roughly 3–5\% over the CoMM backbone. The synergy objective is designed as an inductive bias toward synergistic cross-modal information rather than a strict or provable recovery of the theoretical PID synergy term $  S  $. We view this overhead as a favorable trade-off given the consistent gains in modeling cross-modal interactions. A current limitation of MIIB is its primary focus on static modalities and relatively compact backbones. Future work will extend MIIB to dynamic modalities such as video and explore its integration with large foundation models in multi-source settings, aiming to develop more robust and scalable multimodal AI systems.}



\bibliographystyle{ieeetr}
\bibliography{main}
\begingroup
\makeatletter
\def\@IEEEBIOphotowidth{0.8in}
\def\@IEEEBIOphotodepth{1in}
\def\@IEEEBIOhangwidth{0.94in}
\def\@IEEEBIOhangdepth{1in}
\makeatother
\vspace{-40pt}
\begin{IEEEbiography}[{\includegraphics[width=0.8in,height=1in,clip,keepaspectratio]{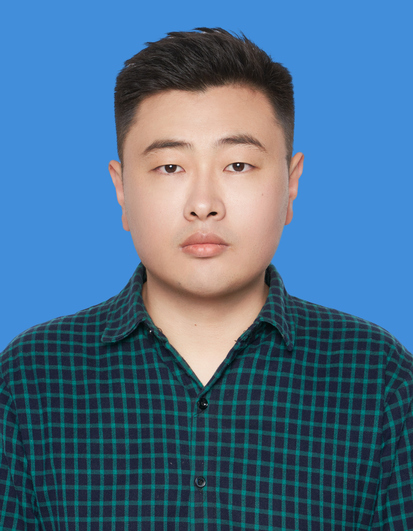}}]{Liangjian Wen} is  currently an associate professor with the School of Computing and Artificial Intelligence,
Southwestern University of Finance and Economics, Chengdu, China.  The Ph.D. degree was obtained from the School of Computer Science and Engineering, University of Electronic Science and Technology of China (UESTC) in July 2021.
His research interests include machine learning, deep learning, representation learning, and self-supervised learning.
\end{IEEEbiography}
\vspace{-42pt}
\begin{IEEEbiography}[{\includegraphics[width=0.8in,height=1in,clip,keepaspectratio]{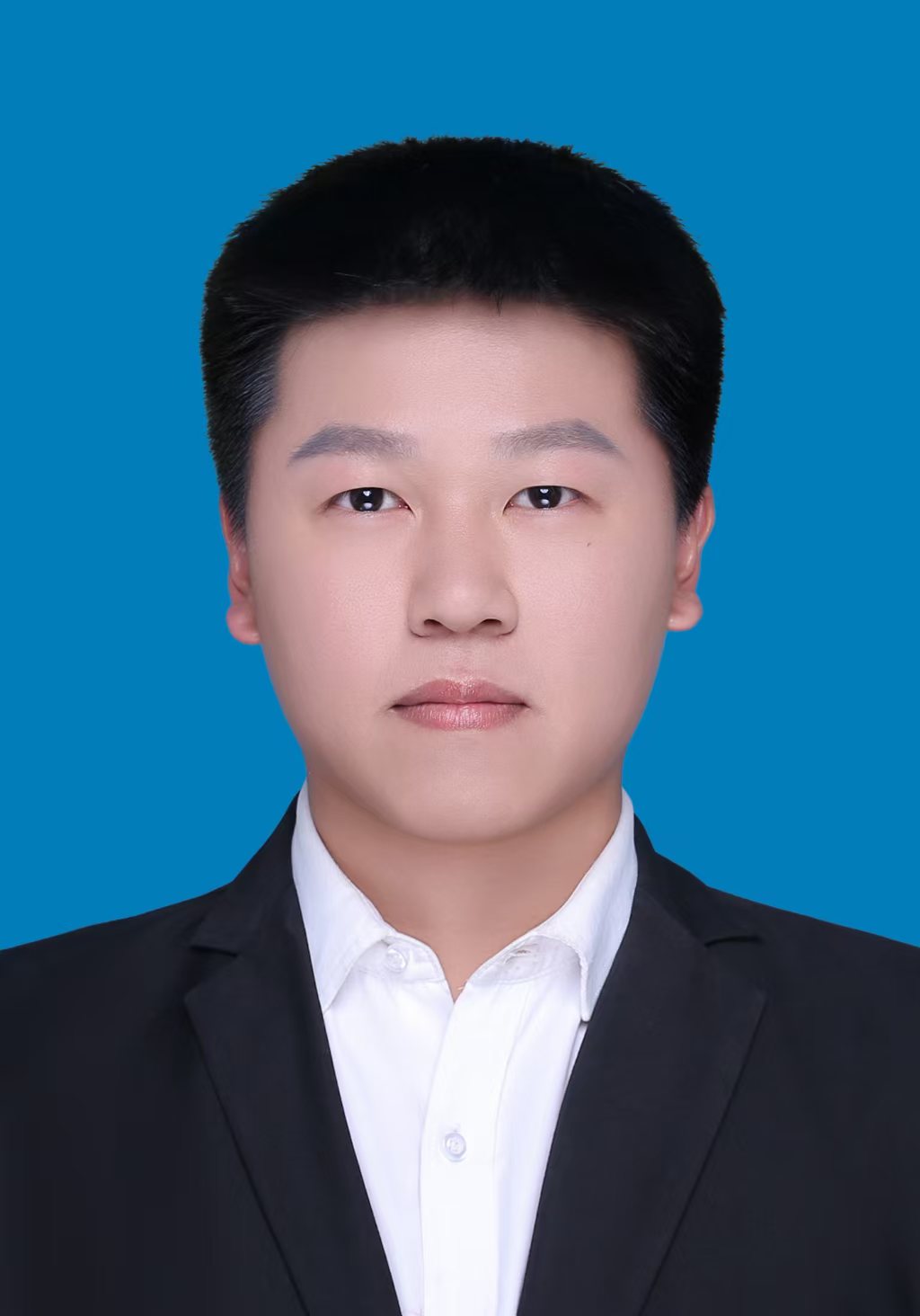}}]{Jiangtao Zheng}
	received the BE degree from Zhengzhou University of Aeronautics, Zhengzhou, China. He is currently pursuing the M.S. degree with the School of Computing and Artificial Intelligence, Southwestern University of Finance and Economics, Chengdu, China.
His research interests include machine learning, deep learning, and representation learning.
\end{IEEEbiography}
\vspace{-42pt}
\begin{IEEEbiography}[{\includegraphics[width=0.8in,height=1in, clip,keepaspectratio]{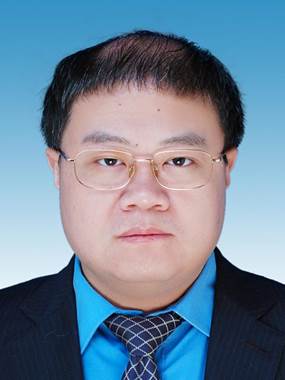}}]{Jiang Duan} received the BSc degree in mechanical engineering from Southwest Jiaotong University, Chengdu, China, in 2001, the MSc degree from the University of Derby, Derby, U.K., in 2002, and the PhD degree from the University of  Nottingham, U.K., in 2006. He is a professor at the Southwestern University of Finance and Economics, China. He is a recognized national expert, the winner of the first Sichuan Outstanding Talent Award (The Highest Talent Award of Sichuan province), the winner of the Sichuan Youth Science and Technology Award, and a standing committee member of the Sichuan Association for Science and Technology. He has published more than 30 academic papers, won more than 10 international and national patents, and his research has been funded by about 20 national and provincial funds.
\end{IEEEbiography}
\vspace{-42pt}
\begin{IEEEbiography}[{\includegraphics[width=0.8in,height=1in,clip,keepaspectratio]{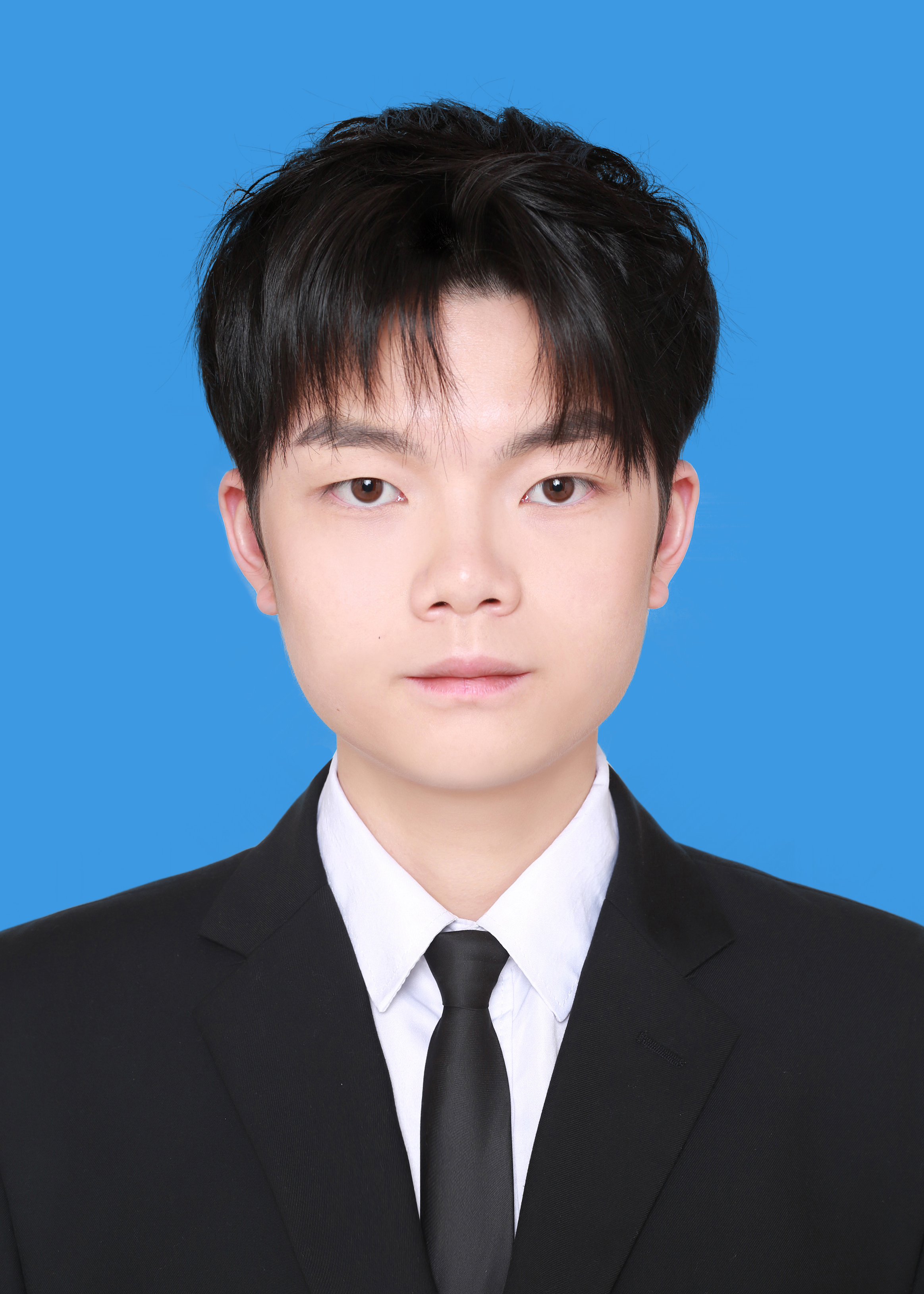}}]{Qun Dai}
	received the BE degree from Southwestern University of Finance and Economics, Chengdu, China, in 2025. He is currently working toward the master's degree with the School of Computing and Artificial Intelligence, Southwestern University of Finance and Economics, Chengdu, China. His research interests include machine learning, deep learning, and representation learning.
\end{IEEEbiography}

\vspace{-42pt}
\begin{IEEEbiography}[{\includegraphics[width=0.8in,height=1in,clip,keepaspectratio]{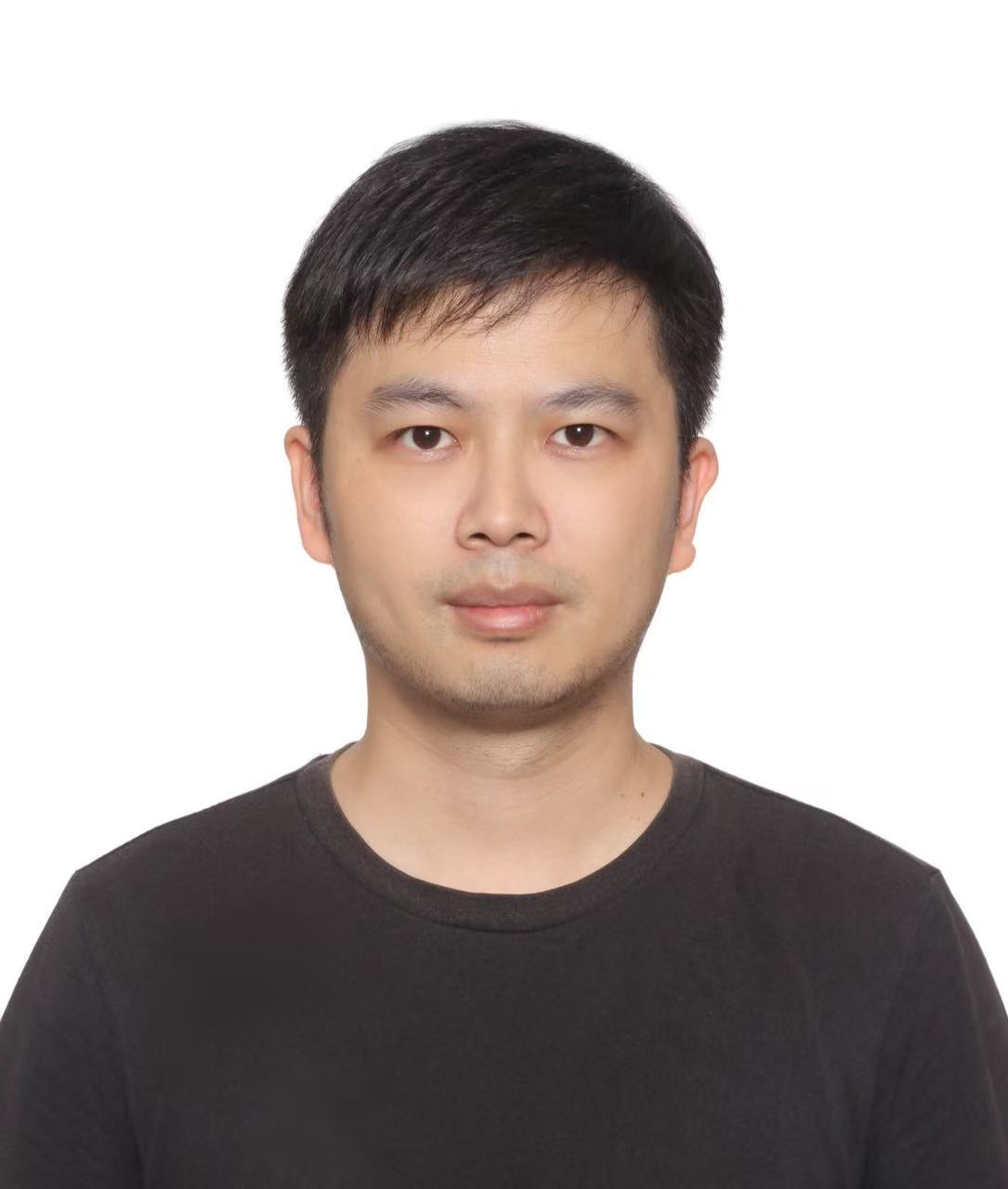}}]{Yong Dai} received the Ph.D. degree from the University of Electronic Science and Technology of China in 2021. He previously conducted research on large language models at Microsoft Research Asia and at Tencent AI Lab. Since May 2025, he has served as Head of Multimodal Large Models at the Beijing Humanoid Robot Innovation Center, leading research on embodied brain-cerebellum systems and world models.
\end{IEEEbiography}
\vspace{-42pt}
\begin{IEEEbiography}[{\includegraphics[width=0.8in,height=1in,clip,keepaspectratio]{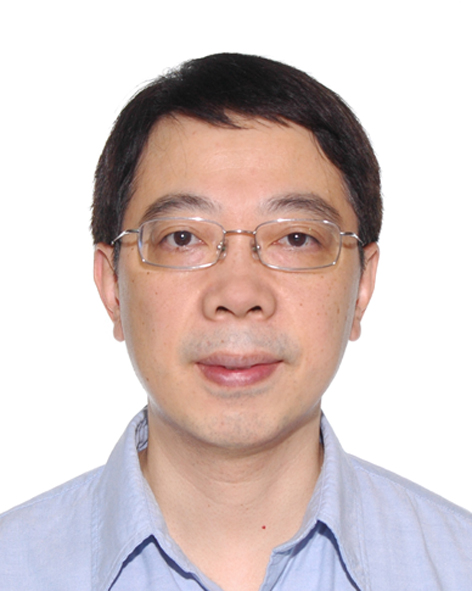}}]{Jianzhuang Liu}  (Senior Member, IEEE) received the PhD degree in computer vision from The Chinese University of Hong Kong, in 1997. From 1998 to 2000, he was a research fellow with Nanyang Technological University, Singapore. From 2000 to 2012, he was a post-doctoral fellow, an assistant professor, and an adjunct associate professor with The Chinese University of Hong Kong, Hong Kong. In 2011, he joined the Shenzhen Institutes of Advanced Technology, Chinese Academy of Sciences, Shenzhen, China, as a professor. He was a principal researcher in Huawei Company from 2012 to 2023. He has authored more than 200 papers in the areas of computer vision, image processing, machine learning, and multimodal AI.
\end{IEEEbiography}
\vspace{-40pt}
\begin{IEEEbiography}[{\includegraphics[width=0.8in,height=1in,clip,keepaspectratio]{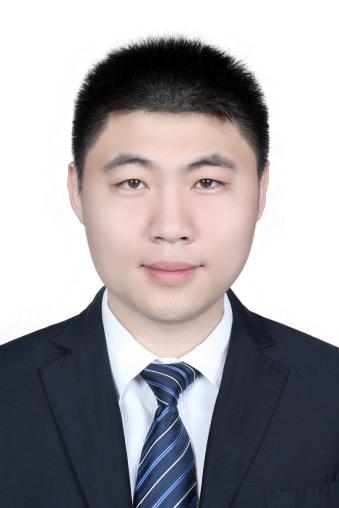}}]{Jun Wang}, Professor, Doctoral Supervisor, Southwestern University of Finance and Economics, School of Management Science and Engineering.  Jun Wang has long been committed to emerging interdisciplinary research in fintech, financial intelligence, and digital-intelligence management. From the perspective of big data analytics, he addresses significant issues in financial markets using advanced machine learning, data analytics, and artificial intelligence methods. 
\end{IEEEbiography}
\vspace{-40pt}
\begin{IEEEbiography}[{\includegraphics[width=0.8in,height=1in,clip,keepaspectratio]{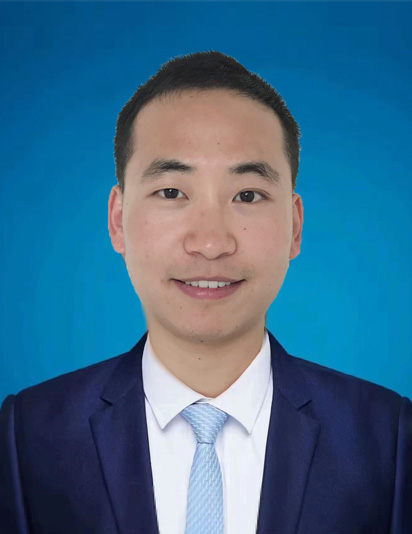}}]{Shu-Dong Huang} (Member, IEEE) received his PhD degree in computer science and engineering from the University of Electronic Science and Technology of China, Chengdu, China. He is currently an associate research professor with the School of Artificial Intelligence, Sichuan University, Chengdu. His research interests include multimodal representation learning, and multimodal Large Language Models.
\end{IEEEbiography}
\vspace{-40pt}
\begin{IEEEbiography}[{\includegraphics[width=0.8in,height=1in,clip,keepaspectratio]{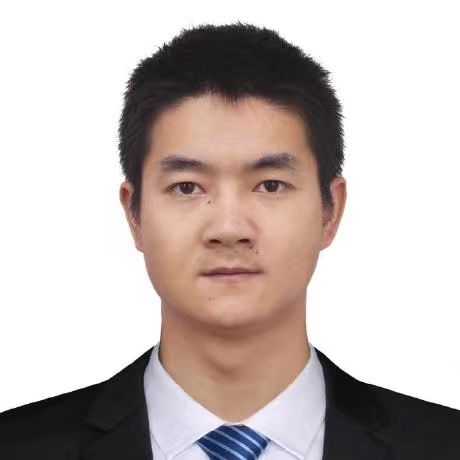}}]{Xinwang Liu} received his PhD degree from National University of Defense Technology (NUDT),
China, in 2013. He is now Professor at School
of Computer, NUDT. His current research interests
include kernel learning, multi-view clustering and
unsupervised feature learning. Dr. Liu has published
80+ peer-reviewed papers, including those in highly
regarded journals and conferences such as IEEE TPAMI, IEEE T-KDE, IEEE T-IP, IEEE T-NNLS,
IEEE T-MM, IEEE T-IFS, ICML, NeurIPS, CVPR,
ICCV, AAAI, IJCAI, etc. He is an Associate Editor
of IEEE T-NNLS and Information Fusion Journal.
\end{IEEEbiography}
\vspace{-40pt}
\begin{IEEEbiography}[{\includegraphics[width=0.8in,height=1in,clip,keepaspectratio]{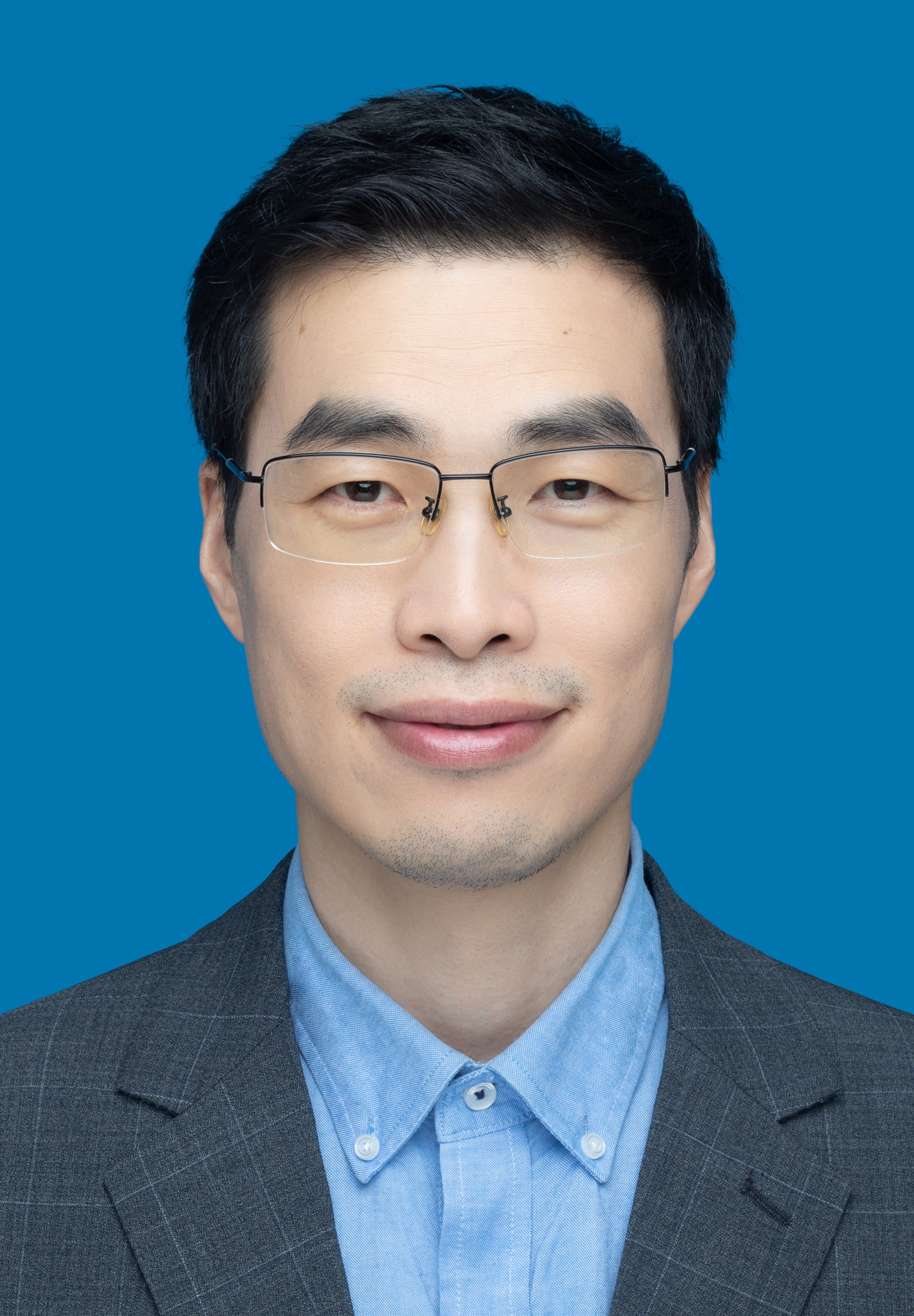}}]{ Zenglin Xu (SM'20)} is a Distinguished Professor at the Artificial Intelligence Innovation and Incubation (AI³) Institute, Fudan University, and an AI Scientist at the Shanghai Academy of AI for Science. He earned his Ph.D. in Computer Science and Engineering from The Chinese University of Hong Kong and completed postdoctoral research at Purdue University and the Max Planck Institute for Informatics. Dr. Xu serves as Vice President of the International Neural Network Society (INNS) and is the Editor-in-Chief of Academia AI and Applications. His research centers on trustworthy machine learning and agentic AI, AI for Science, and domain-specific applications in healthcare, finance and education.
\end{IEEEbiography}
\vspace{-40pt}
\begin{IEEEbiography}[{\includegraphics[width=0.8in,height=1in,clip,keepaspectratio]{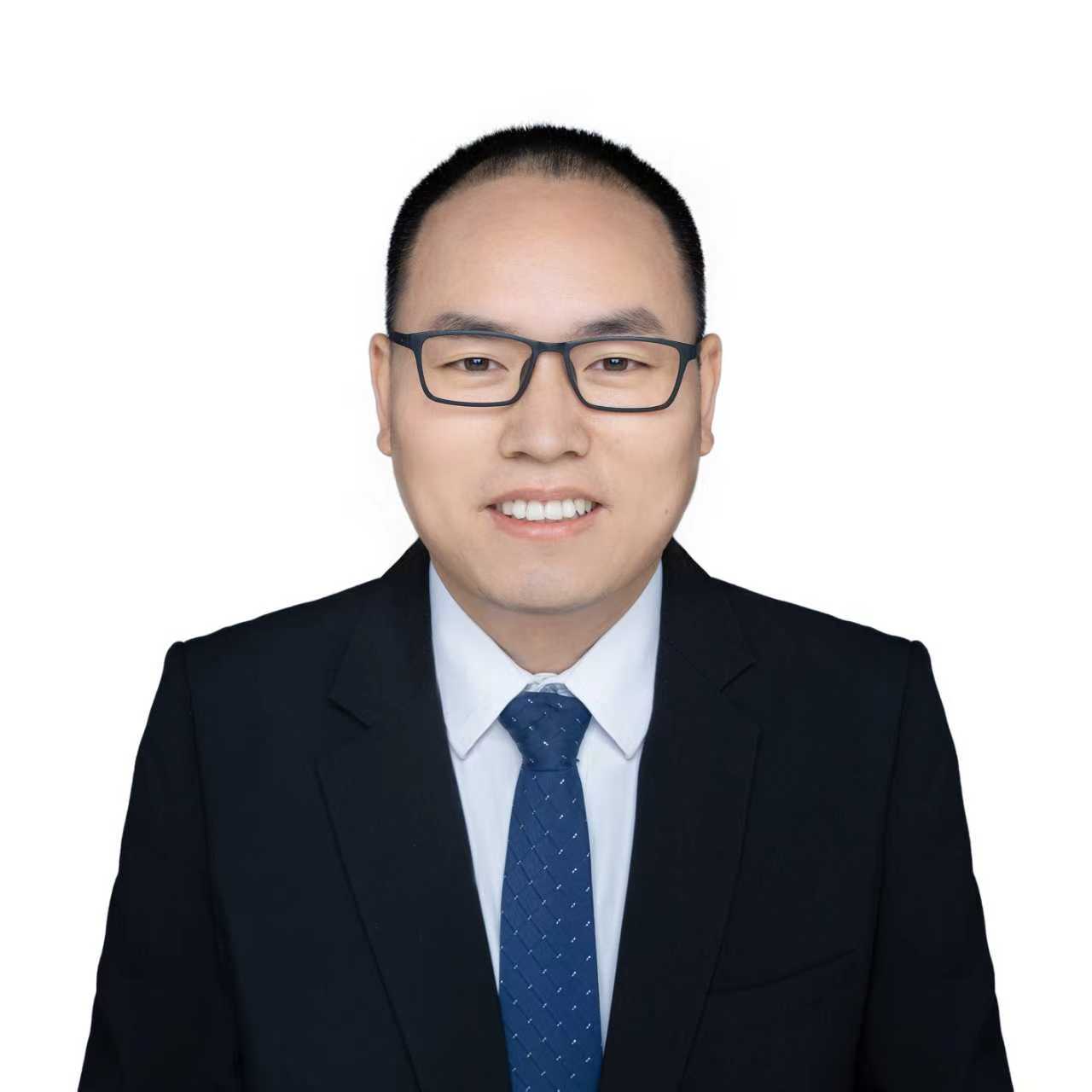}}]{Zhao Kang} received the Ph.D. degree in computer science from the Southern Illinois University Carbondale,
Carbondale, IL, USA, in 2017. He is currently a Professor in
the School of Computer Science and Engineering, University
of Electronic Science and Technology of China, Chengdu,
China.  His research interests include graph machine
learning and large language models. Dr. Kang has been an AC/SPC/PC Member for a number of top conferences, such as NeurIPS, ICML, ICLR, AAAI, IJCAI, CVPR, and SIGKDD. He serves as an Associate Editor for Neural
Networks and Pattern Recognition.
\end{IEEEbiography}
\endgroup

\end{document}